%% file: main.tex
\documentclass[journal]{IEEEtran}
\IEEEoverridecommandlockouts

\usepackage{ctable}
\usepackage{amsmath,amssymb,amsfonts}
\usepackage{graphicx}
\usepackage{textcomp}
\usepackage{xcolor}
\usepackage{stackengine}
\usepackage{scalerel}
\usepackage{graphicx}
\usepackage{multirow}
\usepackage[ruled]{algorithm}
\usepackage{algpseudocode}
\usepackage{lettrine}

\usepackage{xcolor}
\newcommand{\bl}[1]{{\textcolor{blue}{#1}}}
\newcommand{\rl}[1]{{\textcolor{red}{#1}}}

\usepackage[flushleft]{threeparttable}

\usepackage{stfloats}

\usepackage[margin=20pt]{subfig}
\def\BibTeX{{\rm B\kern-.05em{\sc i\kern-.025em b}\kern-.08em
    T\kern-.1667em\lower.7ex\hbox{E}\kern-.125emX}}
\usepackage{soul}

\usepackage[numbers]{natbib}
\usepackage[bookmarks=true]{hyperref}

\usepackage{sparo_acronyms}

\title{\LARGE \bf
	Uni-Mapper: Unified Mapping Framework for Multi-modal LiDARs in Complex and Dynamic  Environments
}

\author{Gilhwan Kang$^{1,2}$, Hogyun Kim$^{2}$, Byunghee Choi$^{2}$, Seokhwan Jeong$^{2}$, Young-Sik Shin$^{3*}$ and Younggun Cho$^{2*}$

\thanks{

This work was supported by Institute of Information \& communications Technology Planning \& Evaluation (IITP) grant (RS-2022-II220448) and National Research Foundation of Korea (NRF) grant (RS-2025-02217000) funded by the Korea government (MSIT) and Korea Agency for Infrastructure Technology Advancement (KAIA) grant funded by the MoLIT (RS-2022-00143717) and National Research Council of Science \& Technology as part of the project titled "Development of Core Technologies for Robot General Purpose Task Artificial Intelligence Framework" (NK254G). \\
*Corresponding authors: Young-Sik Shin and Younggun Cho.
\\
$^{1}$Gilhwan Kang is with the Robotics Lab, Research and Development Division, Hyundai Motor Company, Uiwang, South Korea. ({\tt\small gilhwan@hyundai.com}). \\
$^{2}$Gilhwan Kang, $^{2}$Hogyun Kim, $^{2}$Byunghee Choi, $^{2}$Seokhwan Jeong, and $^{2*}$Younggun Cho are with the Department of Electrical and Computer Engineering, Inha University, South Korea. ({\tt\small [rlfghks527, hg.kim, bhbhchoi, eric5709]@inha.edu, yg.cho@inha.ac.kr}). \\
$^{3*}$Young-Sik Shin is with the Department of AI Machinery, Korea Institute of Machinery and Materials, Daejeon, South Korea. ({\tt\small yshin86@kimm.re.kr}).\\

}%

}

\begin{document}





\input{0_abstract}
\input{1_Intro3}
\input{2_RelatedWorks}
\input{3_SystemOverview}
\input{4_DynamicAwareTriangleDescriptor}
\input{4.5_DynaSTD}
\input{5_MapMerge}

\input{6_EvaluationSetup}

\input{7_EvaluationResults}

\input{9_Conclusion}

\footnotesize
\bibliographystyle{IEEEtranN}

\bibliography{sparo_reference}

\begin{IEEEbiography}[{\includegraphics[width=1in,height=1.25in,clip,keepaspectratio]{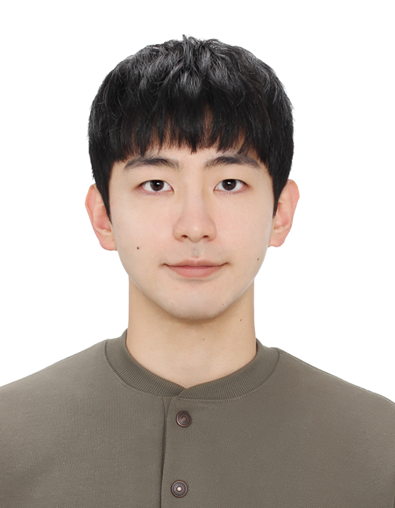}}]{Gilhwan Kang} (S'22) received the B.S. degree in electrical engineering, and the M.S degree in electrical and computer engineering from Inha University, Incheon, South Korea, in 2022 and 2024, respectively. He is currently a research engineer with the Robotics Lab, Research and Development Division, Hyundai Motor Company, South Korea. His research interests include simultaneous localization and mapping (SLAM), long-term map management, and deep learning for 3D scene understanding.
\end{IEEEbiography}
\vspace{-1.2cm}

\begin{IEEEbiography}[{\includegraphics[width=1in,height=1.25in,clip,keepaspectratio]{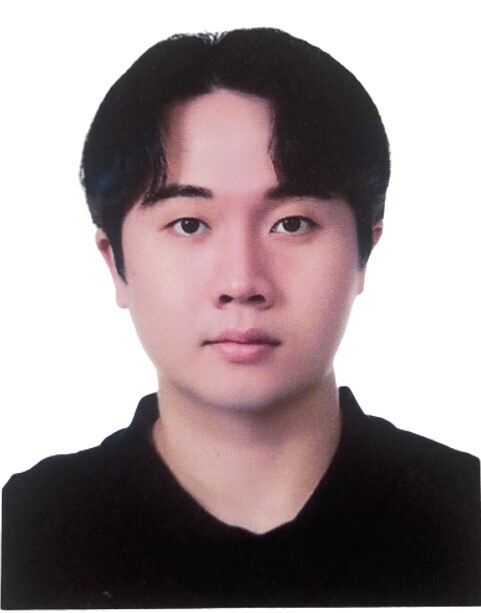}}]{Hogyun Kim} (S'23) received the B.S. degree in naval architecture and ocean engineering from Inha University, Incheon, South Korea, in 2023. He is currently Ph.D. student in the electrical and computer engineering from Inha University. His research interests include simultaneous localization and mapping (SLAM), multi-robot slam, and field robotics.
\end{IEEEbiography}
\vspace{-1.2cm}

\begin{IEEEbiography}[{\includegraphics[width=1in,height=1.25in,clip,keepaspectratio]{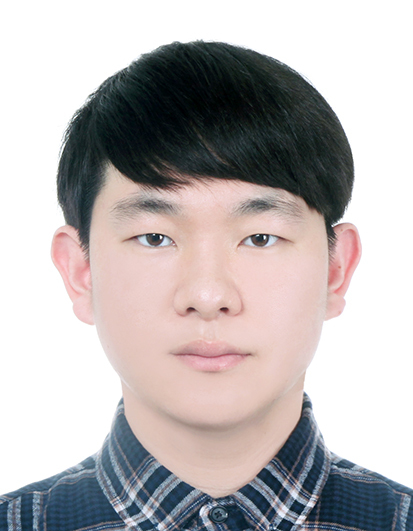}}]{Byunghee Choi} (S'23) received the B.S. degree in the Robotics Engineering from YeungNam University, Gyeongsan, South Korea, in 2023. He is currently M.S. student in electrical and computer engineering from Inha University, Incheon, Korea. His research interests include simultaneous localization and mapping (SLAM), indoor robotics, and radar place recognition.
\end{IEEEbiography}
\vspace{-1.2cm}

\begin{IEEEbiography}[{\includegraphics[width=1in,height=1.25in,clip,keepaspectratio]{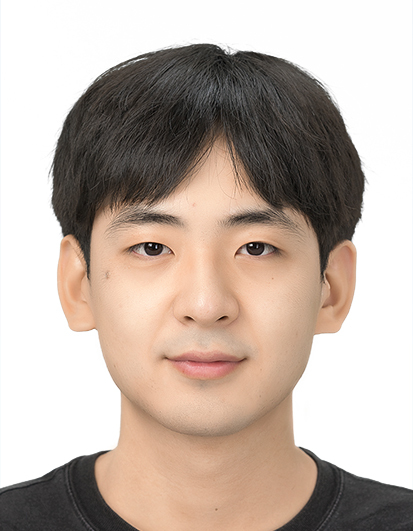}}]{Seokhwan Jeong} (S'23) received the B.S. degree in electronic engineering from from Inha University, Incheon, South Korea, in 2023. He is currently M.S. student in electrical and computer engineering from Inha University. His research interests include simultaneous localization and mapping (SLAM), sensor calibration, and field robotics.
\end{IEEEbiography}
\vspace{-1.2cm}

\begin{IEEEbiography}[{\includegraphics[width=1in,height=1.25in,clip,keepaspectratio]{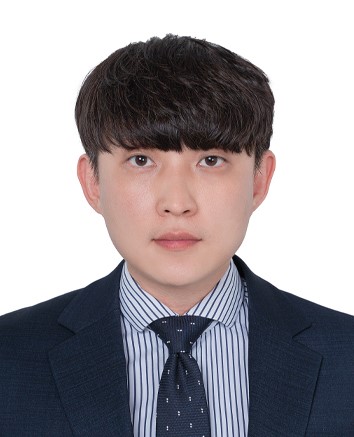}}]{Young-Sik Shin} (S'15--M'20) received the B.S. degree in electrical engineering from Inha University, Incheon, South Korea, in 2013, and the M.S. and Ph.D. degrees in civil and environmental engineering with a dual degree from the Robotics Program, Korea Advanced Institute of Science and Technology (KAIST), Daejeon, South Korea, in 2015 and 2020, respectively. He is currently a Senior Researcher with the Korea Institute of Machinery and Materials (KIMM), Daejeon. His research interests include intelligent robotic autonomy and perception.
\end{IEEEbiography}
\vspace{-1.2cm}

\begin{IEEEbiography}[{\includegraphics[width=1in,height=1.25in,clip,keepaspectratio]{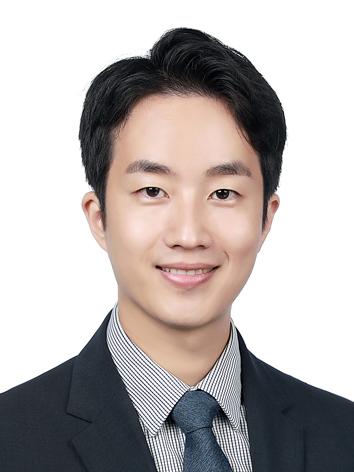}}]{Younggun Cho} (S'15--M'20) received the B.S. degree in electrical engineering from Inha University, Incheon, Korea, in 2013, and the M.S. degree in electrical engineering in 2015 and Ph.D. degree in civil and environmental engineering with dual degree in robotics program (2020) from the Korea Advanced Institute of Science \& Technology (KAIST), Daejeon, Korea. He was an Assistant Professor with the department of Robotics Engineering at Yeungnam University. He is currently an Assistant Professor with the department of Electrical and Electronic Engineering at Inha University, Incheon, South Korea. His current research interests include robust sensing, long-term autonomy, and simultaneous localization and mapping.
\end{IEEEbiography}

\end{document}

%% file: 0_abstract.tex
\maketitle

\begin{abstract}

The unification of disparate maps is crucial for enabling scalable robot operation across multiple sessions and collaborative multi-robot scenarios. 
However, achieving a unified map robust to sensor modalities and dynamic environments remains a challenging problem. 
Variations in LiDAR types and dynamic elements lead to differences in point cloud distribution and scene consistency, hindering reliable descriptor generation and loop closure detection essential for accurate map alignment.
To address these challenges, this paper presents \textit{Uni-Mapper}, a dynamic-aware 3D point cloud map merging framework for multi-modal LiDAR systems. It comprises dynamic object removal, dynamic-aware loop closure, and multi-modal LiDAR map merging modules. A voxel-wise free space hash map is built in a coarse-to-fine manner to identify and reject dynamic objects via temporal occupancy inconsistencies. The removal module is integrated with a LiDAR global descriptor, which encodes preserved static local features to ensure robust place recognition in dynamic environments. 
In the final stage, multiple pose graph optimizations are conducted for both intra-session and inter-map loop closures. We adopt a centralized anchor-node strategy to mitigate intra-session drift errors during map merging. 
In the final stage, centralized anchor-node-based pose graph optimization is performed to address intra- and inter-map loop closures for globally consistent map merging.
Our framework is evaluated on diverse real-world datasets with dynamic objects and heterogeneous LiDARs, showing superior performance in loop detection across sensor modalities, robust mapping in dynamic environments, and accurate multi-map alignment over existing methods. Project Page: \texttt{\url{https://sparolab.github.io/research/uni_mapper}}.


\end{abstract}

%% file: 1_Intro3.tex
\section{Introduction}

\begin{figure}[t]
	\centering
	\def\width{0.8\textwidth}%
    {
    \includegraphics[clip, trim= 0cm 0cm 0cm 0cm, width=\width]{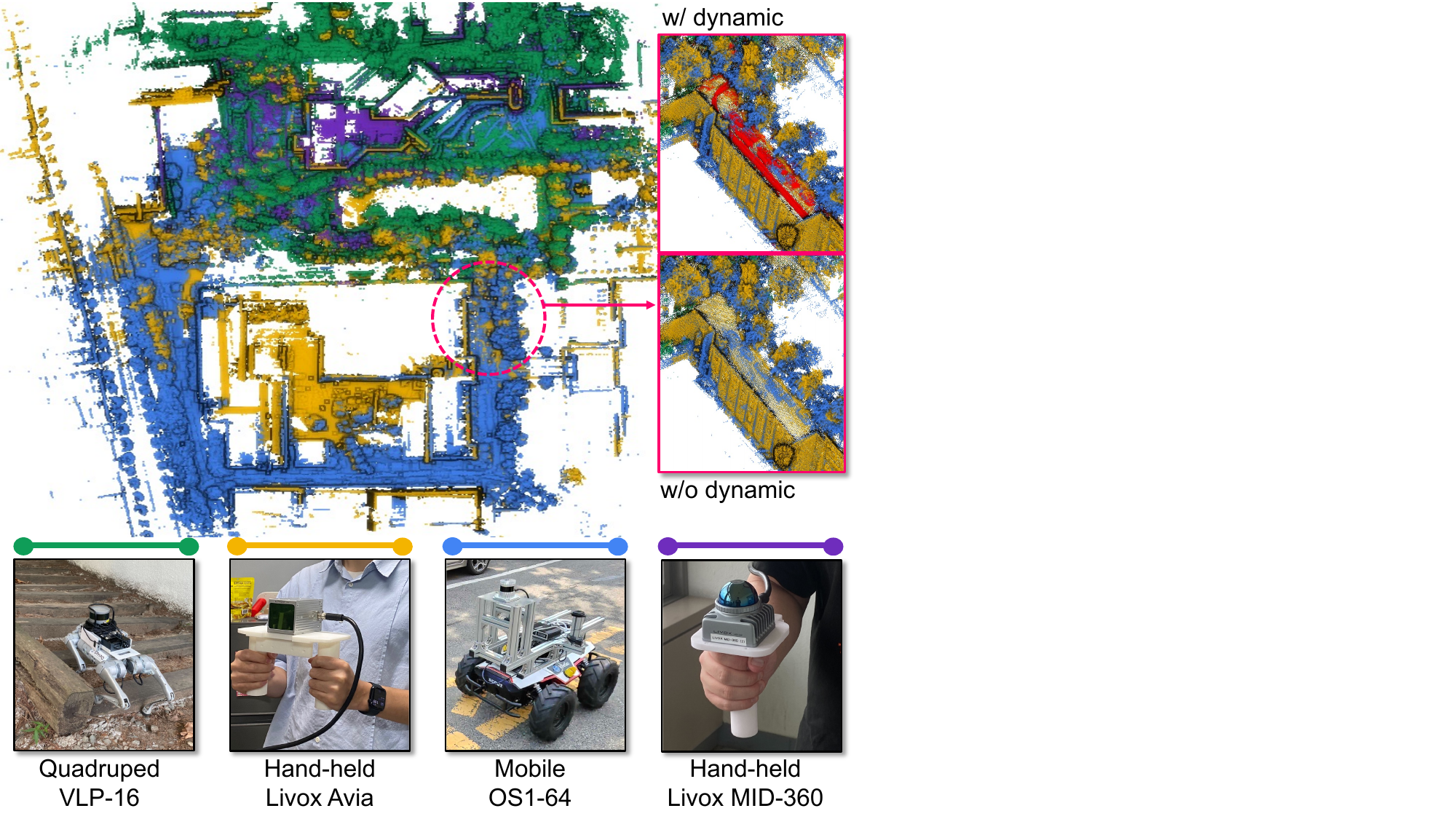}
	}
    \vspace{-0.4cm}
    \captionsetup{font=footnotesize}
    \caption{ The output map from Uni-Mapper. Each data sequence is acquired by various types of LiDARs and platforms. With dynamic-aware and multi-modal LiDAR place recognition, our framework can achieve a static and unified map for different modalities in a single step.}
    
	\label{fig:thumbnail}
\end{figure}

\lettrine[findent=2pt]{\textbf{L}}{arge}-scale mapping \cite{yin2023automerge, liu2023large} serves as the backbone for autonomous navigation systems, providing the spatial data required to operate safely and efficiently in diverse environments. 
Among the various sensors employed for mapping, \ac{LiDAR} stands out as a pivotal sensor due to its exceptional accuracy and comprehensive coverage.
With advancements in sensor technology and the expanding requirements of diverse applications and environments, multiple types of \ac{LiDAR} have emerged, which have distinct \ac{FoV} and scanning patterns. 

Despite the expanded range of LiDAR types enhancing their applicability for diverse platforms, place recognition, which is a crucial technique for integrating multiple maps, is more challenging across LiDAR modalities because of their different data distribution \cite{jung2023helipr, kim2024diter++}.
Current map-merging frameworks \cite{kim2022lt, huang2021disco, zhong2022dcl, kim2025skid} based on conventional place recognition methods \cite{kim2018scan, wang2020lidar, schaupp2019oreos, kim2024narrowing} often lack the capability to effectively merge maps from multi-modal \ac{LiDAR} systems.
Furthermore, dynamic objects (pedestrians, vehicles, bicycles, etc.) are frequently present in the real world. Since the \ac{LiDAR} map is generated by the sequential accumulation of scan data, these dynamic objects cause undesired traces on the map as known as the ghost trail effect \cite{pagad2020robust}, leading to performance degradation in tasks such as place recognition or re-localization for a map-merging process. 
Consequently, effective cross-modality integration techniques and the removal of dynamic objects are imperative for the scalability of multi-agent \ac{SLAM} in diverse real-world environments.

In this paper, we introduce \textbf{Uni-Mapper}, a \ac{LiDAR}-modality-agnostic, dynamic-aware map merging framework. 
Fig. \ref{fig:thumbnail} represents a unified map for multi-robot, multi-modal \ac{LiDAR} mapping systems with estimated dynamic objects (red) on our \textit{INHA} dataset. 
To address the tasks of dynamic object removal and scene description simultaneously, we have employed the common voxel representation for both tasks. The dynamic object removal module applies a coarse-to-fine strategy to effectively estimate free space and remove dynamic objects in real time. Additionally, this module is integrated with \ac{STD} \cite{yuan2023std}, a \ac{SOTA} \ac{LiDAR} descriptor for 3D place recognition. This integration led to an extended descriptor called DynaSTD. 
By leveraging local geometric information from keypoints, DynaSTD is capable of detecting loop pairs across different LiDAR systems and enhancing robustness in dynamic environments by extracting keypoints from static structures, making it suitable for multi-modal \ac{LiDAR} systems and diverse mapping platforms.
In terms of map merging, our approach involves a sequential optimization of both intra-session and multi-map pose graphs. This method employs anchor factors and a two-phase registration process to align multiple disparate maps into a unified coordinate system. 
To summarize, our contributions are as follows:

\begin{itemize}
\item We propose a novel dynamic-aware and LiDAR modality-agnostic map merging framework for multi-session and multi-robot scenarios. Our system simultaneously estimates dynamic points and merges multiple maps in a single process.
\item For efficient dynamic object removal, our method employs a two-level voxel representation that estimates free space via ground segmentation. This representation leverages sliding windows to efficiently estimate unoccupied space, enabling online static mapping.
\item DynaSTD is proposed as a global descriptor integrated with a dynamic removal module, which demonstrates robust performance across LiDAR modalities by leveraging a combination of local descriptors.
\item The proposed framework is evaluated using various types of LiDAR and platforms on public and custom datasets, which include numerous dynamic objects. We demonstrate that our framework shows robust performance in dynamic environments and cross-modal LiDAR sensors.

\end{itemize}


%% file: 2_RelatedWorks.tex
\section{Related Works}

\begin{figure*}[t]
    \centering
	\def\width{0.98\textwidth}%
	\includegraphics[width=\width, trim=0 0 0 0, clip]{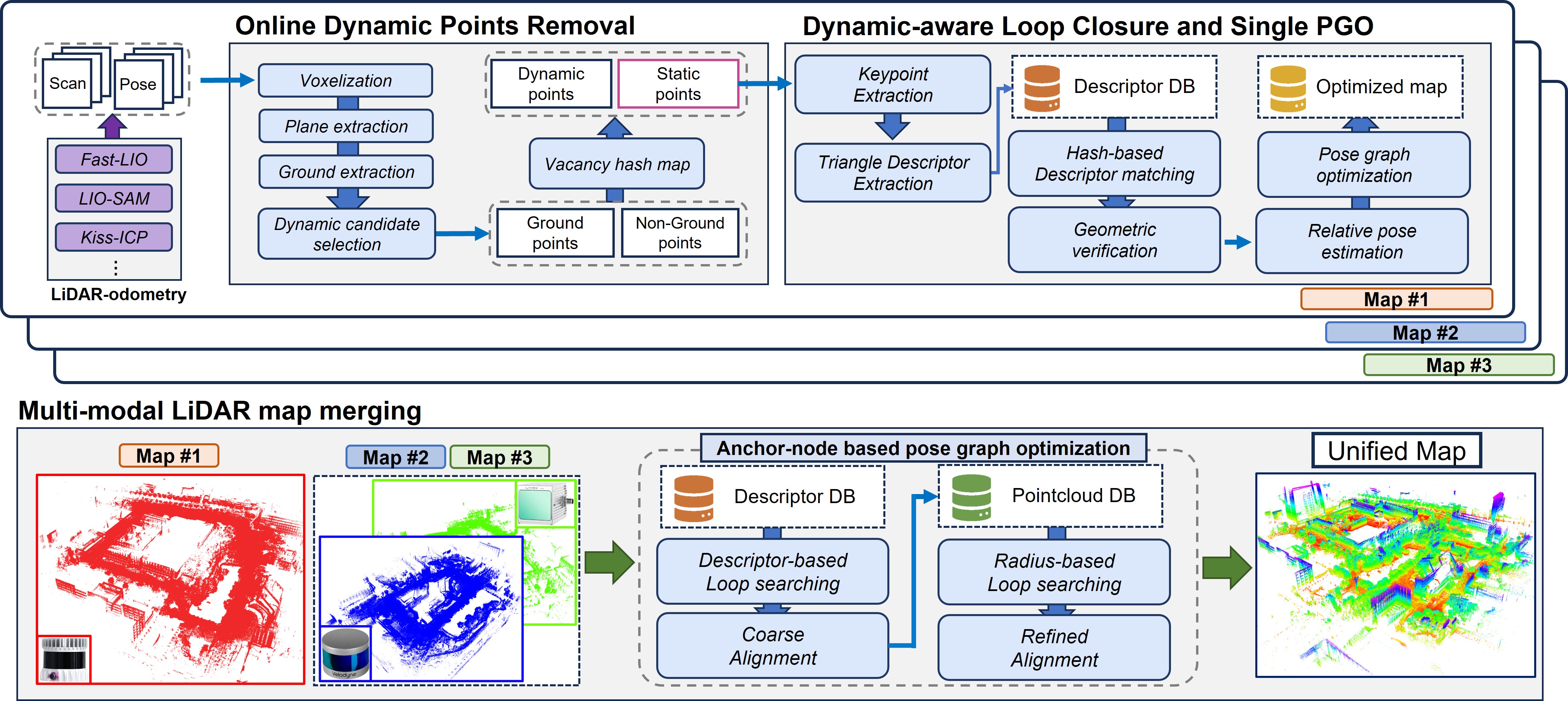}
    \captionsetup{font=footnotesize}
 \caption{Overview of the proposed framework. The framework consists of three main modules: dynamic object removal, dynamic-aware loop closure and multi-modal LiDAR map merging. }
    \label{fig:pipeline}
    \vspace{-0.4cm}
\end{figure*}

\subsection{Dynamic Object Removal}

The \ac{DOR} is a fundamental process for filtering dynamic parts of scans and constructing static maps utilized in robot navigation. 
Existing \ac{DOR} methodologies can be divided into three main streams: i) occupancy-based, ii) discrepancy-based, and iii) learning-based approaches.

Using the ray-tracing algorithm, occupancy-based methods \cite{hornung2013octomap, schauer2018peopleremover} assessed voxel vacancy by updating occupancy probabilities along the traversed path of each ray. The endpoint of the ray is considered occupied, and the traveled voxels are empty. 
Dynablox \cite{schmid2023dynablox} segmented ever-free space based on Truncated Signed Distance Fields (TSDF) by VoxBlox \cite{oleynikova2017voxblox} framework and considered point clouds in the free space as dynamic. However, these methods are computationally expensive when processing an extensive number of \ac{LiDAR} points.
To alleviate computational burden, \citet{duberg2024dufomap} adopted octree structure of UFOMap \cite{duberg2020ufomap} proposed fast ray-casting and classification of void regions considering surrounding voxels based on Chebyshev distance to be robust to sensor noise and localization error.  

Unlike occupancy-based methods, 
discrepancy-based approaches segmented dynamic objects by leveraging the concept that dynamic objects are transient and their measurements exhibit temporal inconsistencies.
Visibility-based methods \cite{fpomerleau2014trail, yoon2019mapless, fan2022dynamicfilter, jpark2022dr} detect temporal inconsistencies in projected range-images, offering computational advantages over occupancy-based methods, especially in dynamic environments. 

However, the incidence angle ambiguity problem causes misclassification of dynamic objects which is far from the sensor.
\citet{kim2020remove} alleviated this issue by using a multi-resolution range image and reverting approach to recover falsely removed static points. 
\citet{wu2024moving} proposed an M-detector, which categorizes occlusions caused by dynamic objects into three cases and detects occluded points at the pixel level in a depth image. To enhance performance, they employed clustering and region-growing techniques. 

Unlike visibility-based methods, ERASOR \cite{lim2021erasor} proposed a 2.5D pseudo-occupancy descriptor based on ground segmentation and classify dynamic points by scan-to-map descriptor differences in an offline manner.
BeautyMap \cite{jia2024beautymap} encoded the map into a 3D binary matrix and compared it with the scan measurement for efficient dynamic removal.

Recently, learning-based approaches have shown significant improvement in moving object segmentation. 
These approaches adopted 2D convolution on range image \cite{chen2021moving,kim2022rvmos} or sparse convolution on 3D voxels \cite{mersch2022receding, mersch2023building} for moving object segmentation. 
While learning-based methods can segment dynamic points in a single scan, they are not suitable for various types of \ac{LiDAR} they are not trained on and require GPU for real-time performance.

\subsection{Place Recognition}

In multi-map scenarios, \ac{PR} is a key issue for a map merging process. 
Depending on the approach to solve the problem, it can be categorized as follows: i) local descriptors based on point features, ii) global descriptors by structural context, and iii) learning-based approaches.

\citet{shan2021imaging} proposed a \ac{PR} method that uses ORB features and the DBoW \cite{glopez2012dbow} for high-resolution LiDAR intensity images. They also introduced an efficient and robust outlier rejection technique based on PnP RANSAC. \citet{mbosse2013pr} performed keypoint extraction directly on 3D data. They then created the Gestalt Descriptor, which represents the surrounding information for each keypoint. However, these keypoint-based approaches are sensitive to the density and noise of the measurements.

For robust \ac{PR}, global descriptors are preferred utilizing structural information from the scene. M2DP \cite{he2016m2dp} and Scan Context (SC) \cite{kim2018scan} encoded 3D spatial information into a low-dimensional space and extracted geometric features, which leads to information loss. There were several methods \cite{luo2023bevplace, xu2023ring++} that create BEV images through projection and utilize these images for place recognition. However, these projection-based methods are relatively vulnerable to significant viewpoint changes.
STD \cite{yuan2023std} combined the advantages of local and global descriptors by exploiting a global representation built from local triangular keypoint features, achieving 6-DOF pose invariance. However, dynamic objects prevalent in real-world environments can be regarded as inconsistent features for \ac{PR}.

Recently, learning-based methods \cite{uy2018pointnetvlad, ma2022overlaptransformer, vidanapathirana2022logg3d} have been to the forefront, particularly excelling in scenarios where large datasets are available for training. 
OverlapTransformer (OT) \cite{ma2022overlaptransformer} used range-image from point cloud and embedded a rotation-invariant global descriptor using a transformer architecture. LoGG3D-Net \cite{vidanapathirana2022logg3d} leveraged local consistency loss to guide the network in capturing local features and ensuring consistency to revisit.  
While learning-based methods show satisfactory performance on trained datasets, they struggle to generalize across multi-modal sensors and multi-robot platforms due to their dependency on training datasets.

\subsection{Multi-Map Merging}

Multi-map merging, including multi-session and multi-robot collaborative mapping, is a critical issue for long-term map maintenance and incremental map expansion. 

Recent research \cite{liu2023large, liu2023efficient} adopted bundle adjustment to LiDAR point cloud map using geometric plane and line features. Although these methods can improve both single and multi-session map consistency, they depend  on prior relative pose for long-term data association. 
To tackle this issue, \citet{yu2023multi} proposed a lightweight multi-session mapping framework with global data association by utilizing the affine Grassmannian manifold \cite{lusk2022graffmatch}.

Without any prior information, the core of most map merging systems is to identify loop pairs across multiple maps and apply pose graph optimization \cite{dellaert2017factor}.
DiSCo-SLAM \cite{huang2021disco} and DCL-SLAM \cite{zhong2022dcl} utilized lightweight descriptors \cite{kim2018scan, wang2020lidar} to achieve real-time communication for homogeneous robot platform and sensor configuration. LAMP 2.0 \cite{chang2022lamp}, Swarm-SLAM \cite{lajoie2023swarm}, and FRAME \cite{stathoulopoulos2023frame} proposed collaborative mapping systems for heterogeneous robot swarm. However, omnidirectional descriptors \cite{kim2018scan, wang2020lidar, schaupp2019oreos} utilized in these frameworks are difficult to detect loop pairs with different types of \ac{LiDAR} because of their dependency on \ac{FoV}.

Unlike the aforementioned studies, which concentrate on multi-robot mapping, LTA-OM \cite{zou2024lta} proposed a multi-session localization and mapping framework that performs real-time localization on a prior map while maintaining map consistency with the existing map through long-term association.
However, these methods do not account for dynamic objects, which can result in loop detection failures in complex dynamic environments.

LT-mapper \cite{kim2022lt} aligned multiple maps from different sessions based on a modular pipeline containing \ac{SLAM} \cite{kim2022sc} and \ac{DOR} \cite{kim2020remove}.
However, LT-mapper only operates with homogeneous \ac{LiDAR}, and dynamic objects are post-processed to accommodate map changes for each session.

Our proposed mapping framework, \textbf{Uni-Mapper}, unlike the aforementioned methods, is suitable for multi-modal LiDARs and platforms with an online \ac{DOR} module, which can be applied to multi-robot rendezvous scenarios in complex environments.

%% file: 3_SystemOverview.tex
\section{System Overview}

The proposed system is built upon three main modules: online dynamic object removal, dynamic-aware loop closure, and multi-modal LiDAR map merging, as shown in Fig. \ref{fig:pipeline}. 
Regardless of the front-end \ac{LiDAR} odometry algorithms \cite{xu2021fast, shan2020lio,vizzo2023kiss,cho2019deeplo}, each module is conducted sequentially when \ac{LiDAR} scans and poses of multiple maps are provided.

First, the voxel grid is used as a base representation. After voxel-wise point cloud analysis, ground areas are estimated. 
We utilize ground information as a prior for free space estimation, employing a two-level hash map composed of coarse and fine voxels. Free space is determined from ground heights to the minimum heights of non-ground voxels using a coarse-to-fine approach. This method, leveraging temporal discrepancies in the free space hash map, allows for the efficient and precise removal of dynamic entities from the scene.

For scene description, STD \cite{yuan2023std} is adopted as the baseline descriptor. This baseline is enhanced by incorporating an online dynamic removal module, resulting in a dynamic-aware scene descriptor, referred to as DynaSTD. It enhances the discriminability of the descriptor in dynamic environments. As DynaSTD encapsulates geometric features of the local scene, it enables effective place recognition across multi-modal \ac{LiDAR} systems. 

Finally, the system implements intra-session loop closure and centralized multiple map merging. 
Anchor-node-based pose graph optimization is adopted to reduce the remaining drift error of single-session odometry in the map merging process. Through an initial and refined registration process, multiple maps are aligned in a unified coordinate system.
Our framework facilitates the merging of multiple maps with an online \ac{DOR} in a single process, making it suitable for both offline map maintenance and centralized multi-agent SLAM.

%% file: 4_DynamicAwareTriangleDescriptor.tex
\section{Online dynamic object removal} \label{sec:1}

Our online dynamic removal module is designed to efficiently reject outliers and ensure robust place recognition in dynamic environments. First, the preprocessing module converts the scans from the LiDAR odometry method to a voxelized format and extracts geometric primitives. Then, the main dynamic removal module is applied to efficiently separate dynamic and static point clouds online. This module consists of three submodules: ground-assisted dynamic candidate voxel selection, two-level hash map-based coarse-to-fine free space estimation, and online dynamic point cloud segmentation.

\subsection{Preprocessing}   \label{subsec:3_A}

From a \ac{LiDAR}-based odometry algorithm, \ac{LiDAR} scans are represented in world coordinates by the estimated $\mathrm{SE}(3)$ poses for each frame. We accumulate every $N_a$ scans to generate keyframe dense point cloud $\mathbf{P}_{k} = \left\{\mathbf{p}_i\right\}_{i=1}^n$, where point $\mathbf{p}=\left(x, y, z\right) \in \mathbb{R}^3$ at keyframe index $k$. For the sake of brevity, the subscript $k$ will be omitted in the following sections. $\mathbf{P}$ is divided into a voxel grid with a pre-defined coarse leaf size $L_c$(e.g., 2m). We define $\mathcal{V}$ as a set of voxels and all points are allocated to the voxel index set $\mathcal{I}$ as,

\begin{equation}
\label{eq:voxelize}
\mathcal{I}=\left\{{I} \mid {I} = (I_x,I_y,I_z) = \left\lfloor\frac{1}{L_c}\left(x, y, z\right)^T\right\rfloor \in \mathbb{Z}^3 \right\}.
\end{equation}
A voxel element ${V}_I \in \mathcal{V}$ contains point cloud subset $\mathbf{P}_I \subset \mathbf{P}$. For the analysis of voxel-wise geometric structure, the distribution of $\mathbf{P}_I$ is extracted as :
\begin{equation}
\label{eq:PCA}
\overline{\mathbf{p}}_I=\frac{1}{n_I} \sum_{j=1}^{n_I} \mathbf{p}_j , \quad \boldsymbol{\Sigma}_I=\frac{1}{n_I} \sum_{j=1}^{n_I}\left(\mathbf{p}_j-\overline{\mathbf{p}}_I\right)\left(\mathbf{p}_j-\overline{\mathbf{p}}_I\right)^T,
\end{equation}
where $n_I$ is the total number of points in $V_I$. The central point and covariance matrix of $\mathbf{P}_I$ are denoted as  $\overline{\mathbf{p}}_I$ and $\boldsymbol{\Sigma}_I$, respectively. To investigate the planarity of each voxel, \ac{PCA} \cite{abdi2010principal} is employed and if the smallest eigenvalue $\lambda_s$ is less than a certain threshold, it is considered to be an element of plane voxel index set $\mathcal{P} \subset \mathcal{I}$, and the eigenvector corresponding to $\lambda_s$ is referred to as the plane normal $\mathbf{n}_I$.

\subsection{Dynamic candidate selection}
\begin{figure}[t]
	\centering
	\def\width{1.0\columnwidth}%
    {
    \includegraphics[clip, trim= 0cm 0cm 0cm 0cm, width=\width]{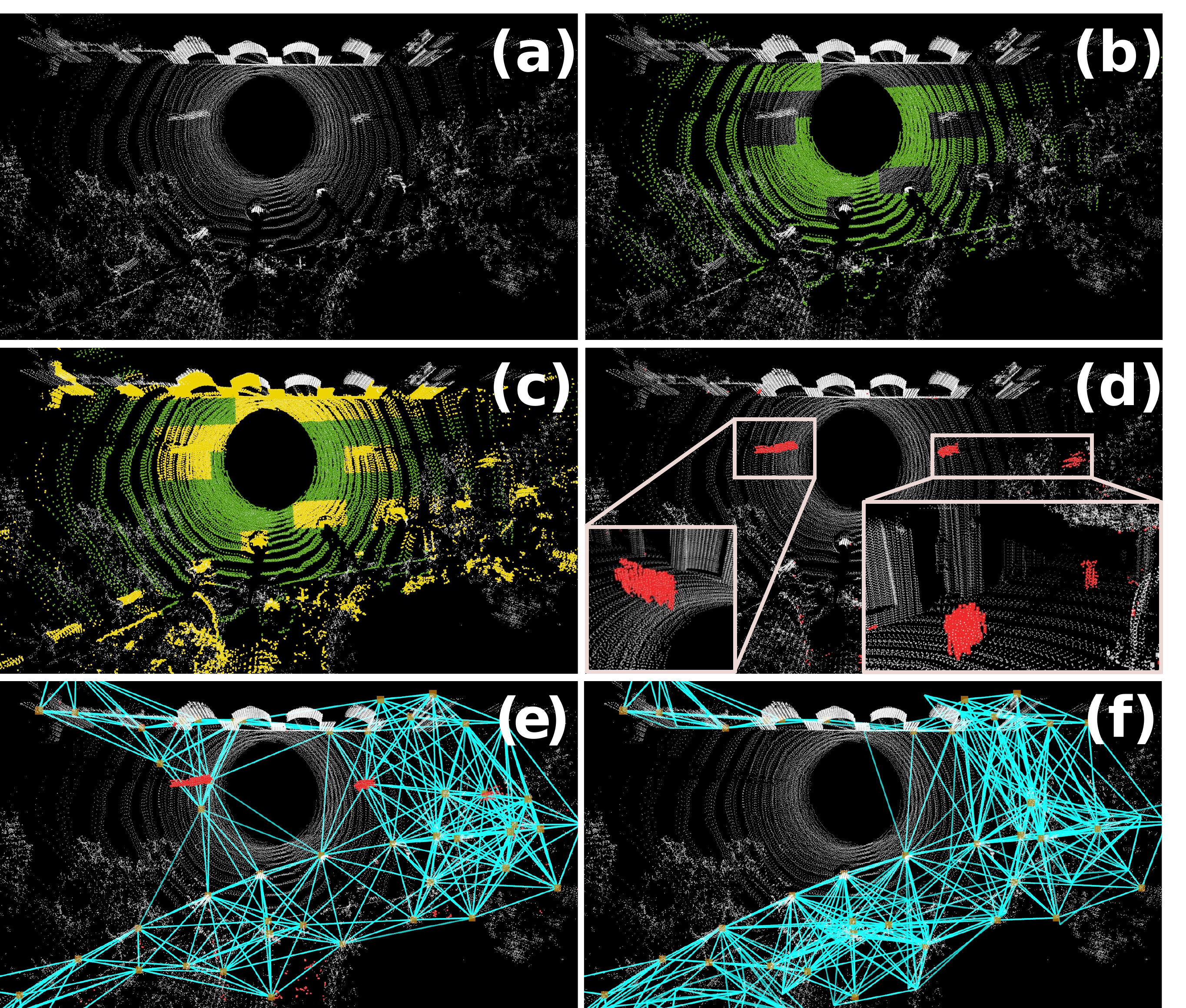}
	}
    \captionsetup{font=footnotesize}
    \vspace{-0.3cm}
    \caption{
    Four steps of the dynamic object removal module (a-d) and comparison of the triangle scene descriptor (e-f) on the \textit{INHA} dataset. (a) Accumulated keyframe point cloud including three dynamic persons represented in world coordinates. (b) Ground voxel extraction only including ground points (green). (c) Dynamic candidate voxel selection where the trace of dynamics can be navigated (yellow). (d) Segmented dynamic points using the online dynamic removal module (red). (e) Constructed local triangle descriptors (turquoise) with dynamic objects and (f) without dynamics.
    }
\label{fig:dyna_module2}
\vspace{-0.4cm}
\end{figure}

The point cloud $\mathbf{P}$ can be divided into two categories: ground points $\mathbf{P}^g$ and non-ground points $\mathbf{P}^{ng}$. As reported by \citet{lim2023erasor2}, most dynamic point clouds such as traces of pedestrians or terrestrial vehicles are classified as $\mathbf{P}^{ng}$ situated on the ground. 
Based on this assumption, candidate locations for dynamic objects can be defined as adjacent regions to ground areas. As shown in Fig. \ref{fig:dyna_module2}, we denote a set of ground voxel indices as $\mathcal{G} \subset \mathcal{P}$, which only contains $\mathbf{P}^g$ and the dynamic candidate voxel index set $\mathcal{C}$ as voxel indices neighboring to $\mathcal{G}$.
Let $C \in \mathcal{C}$ be a candidate voxel element and $\mathbf{P}_C$ be a point cloud in $V_C$.

First, $\mathcal{G}$ is selected where the relative angle between the plane normal $\mathbf{n}_I$ calculated in Sec. \ref{subsec:3_A} and robot uprightness is less than angle margin $\theta_{A}$ as follows: 
\begin{equation}
\label{eq:4}
\mathcal{G}=\left\{{I} \mid I \in \mathcal{P} \wedge \frac{\mathbf{n}_I \cdot \mathbf{z}}{\left\|\mathbf{n}_I\right\| \cdot \left\|\mathbf{z}\right\|} > cos(\theta_{A})\right\},
\end{equation}
where $\theta_{A}$ is set to $30^\circ$ and $\mathbf{z}$ describes the z-axis of the sensor coordinate. 
However, there may be false positives in which the roof or leaf of the tree is also regarded as ground.
As precise ground estimation is essential for dynamic candidate selection, ground voxels that have points below are rejected.

After estimating ground voxels, $\mathbf{P}^{g}_C$ and $\mathbf{P}^{ng}_C$ are segmented using the ground plane equation of $V_C$.
Considering the structural continuity of the ground, the initial plane coefficients of $V_C$ can be inferred from nearby ground voxels.
Let $\mathbf{n}_{a}$ and $\mathbf{p}_{a}$ denote the average plane normal and average center point of the neighboring ground voxels, respectively.

$\mathbf{P}^{g}_C$ is segmented as below:
\begin{equation}
\mathbf{P}^{g}_C=\left\{\mathbf{p} \in \mathbf{P}_C \mid \mathbf{n}_{a}^{T}(\mathbf{p}-\mathbf{p}_a)<m_{th}\right\}.
\end{equation}
We define $m_{th}$ as the ground margin threshold and set it as $0.1m$.

\subsection{Coarse-to-fine free space estimation} \label{subsec:4_B}

Free space is defined as a non-occupied area that is navigable for dynamic objects. Upon this definition, transient $\mathbf{P}^{ng}$ detected within free space can be classified as dynamic objects. 

As the large leaf size of $L_c$ is improper for detailed free space estimation, we subdivide $V_C$ where dynamic objects can exist into finer voxels, akin to the process in Eq. (\ref{eq:voxelize}), using a smaller leaf size $L_f$ (e.g., 0.2m). For clarity, $I$ in Eq. (\ref{eq:voxelize}) is specified as the coarse voxel index $I^c$, and we define the fine voxel index as $I^f$. For the indices of the entire voxel map, we refer to the combined coarse-to-fine voxel indices as $I^{cf}$.

\begin{figure}[t]
	\centering
	\def\width{1.1\columnwidth}%
    {
    \includegraphics[clip, trim= 2cm 2cm 0cm 2cm, width=\width]{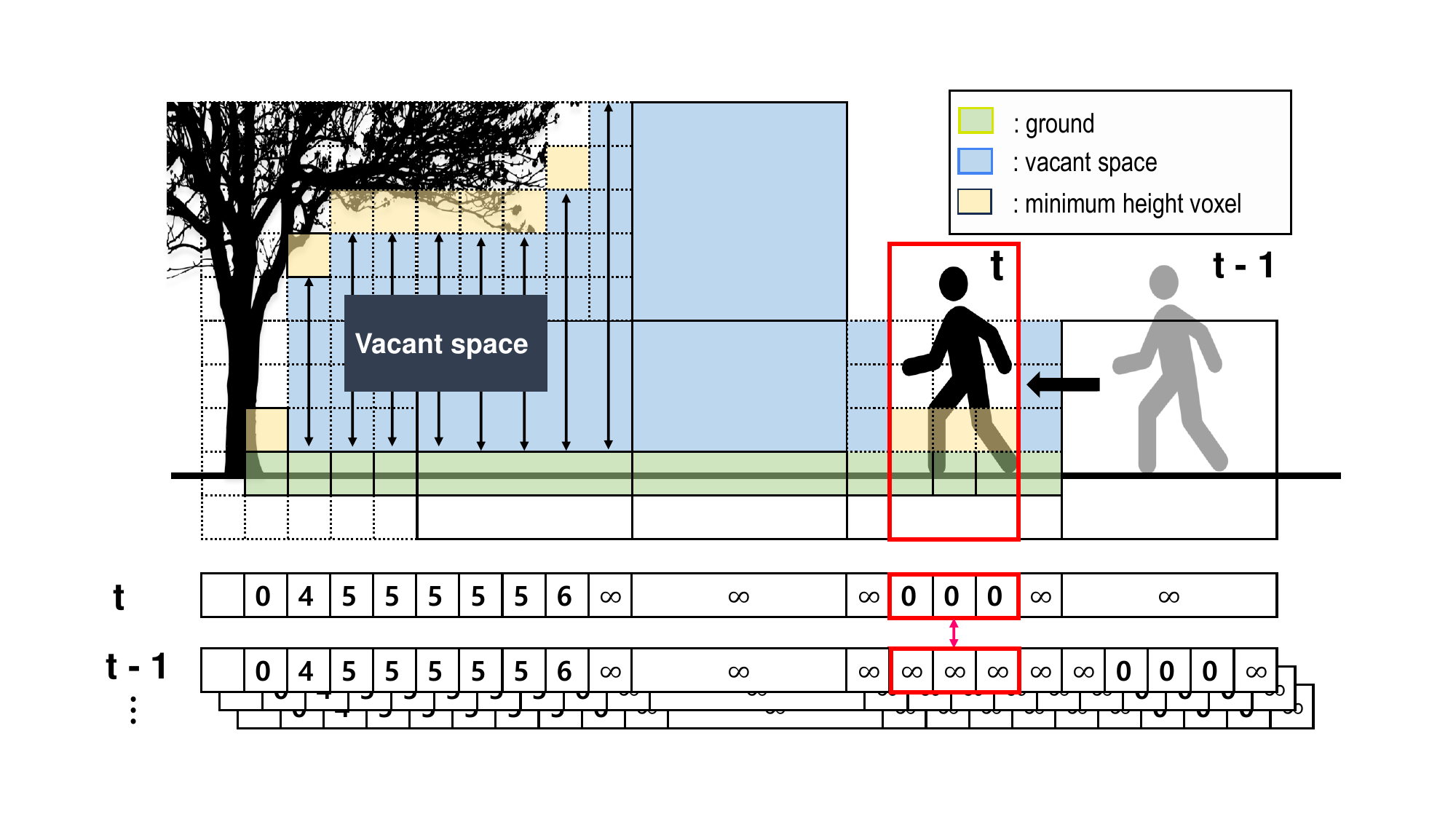}
	}
    \captionsetup{font=footnotesize}
    \vspace{-0.4cm}
    \caption{Visual illustration of the temporal differences in the free space voxel map. For a simplified explanation, the voxel map is projected in a 1D form. When there are no minimum height voxels, the minimum height is represented as infinity.}
	\label{fig:method}
\end{figure}

Instead of relying on ray tracing-based occupancy updates~\cite{hornung2013octomap}, which are computationally expensive, we define the subset of the voxel map that is used to efficiently segment the spatial area:
the free space voxel index set $\mathcal{I}^{cf}_{(I_x,I_y)}$, which is consists of the empty voxels from the ground to the nearest non-ground voxel in the z-axis direction. $\mathcal{I}^{cf}_{(I_x,I_y)}$ is represented as follows:
\begin{equation}
\mathcal{I}_{\left(\mathcal{I}_x, \mathcal{I}_y\right)}^{c f}=\left\{I^{c f} \mid I^{c f}=\left(I_x, I_y, I_z\right), \quad z^g < I_z < z^m\right\},
\end{equation}
where $z^g, z^m \in \mathbb{Z}^1$ are the z-axis index of the ground and the lowest non-ground voxel, respectively. In summary, $\mathcal{I}^{cf}_{(I_x,I_y)}$ represents the set of voxel indices that share the same 2D coordinate on the xy plane. As shown in Fig. \ref{fig:method}, the free space estimation problem is simplified by selecting vacant space between $z^g$ and $z^m$ with a coarse-to-fine approach, which enables real-time free space segmentation. 

As region-wise $z$ indices of vacant space are defined, we denote the free space voxel map $\mathcal{E}$ as follows:
\begin{equation}
\mathcal{E}=\left\{I_{\left(I_x, I_y\right)}^{cf} \mid\left(I_x, I_y\right) \in\{\mathcal{C} \cup \mathcal{G}\}\right\}.
\end{equation}
where $\mathcal{E}$ is all voxels of the current keyframe that lie between the ground and the lowest non-ground voxels. 
Since free space estimation only utilizes $\mathcal{C}$ and $\mathcal{G}$ in a coarse-to-fine manner, this process can be handled efficiently.

\subsection{Online dynamic point cloud segmentation}

For dynamic points segmentation, we leverage sliding windows of sequential measurement and adopt a recursive binary Bayes filter \cite{mersch2022receding}. Let log-odds function $l(x) = \log \frac{p(x)}{1-p(x)} $. Recursive update of dynamic voxels is as follows:  

\begin{equation}
\small
l(m_{i}\mid\mathcal{Z}_{0:t}) = 
\begin{cases} 
l(m_{i}\mid\mathcal{Z}_{0:t-1}) + l(m_{i} \mid \mathcal{Z}_{t}) - l(m_{i}), & \text{if } i \in \mathcal{Z}, \\
l(m_{i} \mid \mathcal{Z}_{0:t-1}), & \text{otherwise},
\end{cases}
\label{eq:bf}
\end{equation}
where $m_i = \{0,1\}$ is a binary dynamic state of fine voxels and $\mathcal{Z}_t=\{ \mathcal{I},\mathcal{E}\}$ is a voxel hash map at timestamp $t$.
As shown in Algorithm \ref{alg:algo1}, fine voxels in current candidate voxels $I_C^f$ are compared with previous $N$ frames of $\mathcal{Z}$ as in Eq. \ref{eq:bf}. 

Update condition of $m$ is as belows:
\begin{equation}
m = 
\begin{cases} 
0 \text{ (static)}, & \text{if } I_C^f \in \mathcal{I} \\
1 \text{ (dynamic)}, & \text{elseif } I_C^f \in \mathcal{E}. 
\end{cases}
\label{eq:dynamic_update}
\end{equation}

In case the voxel of $I^{f}_C$ is not measured in $\mathcal{Z}$, the dynamic state remains unchanged.
Finally, if the vacant probability of $I^{f}_C$, $p(x)=\log \frac{l(x)}{1+exp(-l(x))}$ is larger than 0.5, which indicates this region is not consistently occupied region, $\mathbf{P}^{ng}_C$ in a $I^{f}_C$ is regarded as dynamic points.

\begin{algorithm}[t]
\caption{Online Dynamic Removal}
\label{alg:algo1}
\begin{algorithmic}[1]
\State \textbf{Input:} current frame candidate voxel $^{t}\mathcal{C}$, sliding window of free space hash map $\mathbb{Z}=[^{t-n}\mathcal{Z},..., ^{t-1}\mathcal{Z}]$ 
\State \textbf{Output:} dynamic point cloud $\mathcal{P}^{d}$
\State Let $p(I)$ a dynamic probability of voxel $I$ 
\State \# Check all candidate voxels in the current frame
\For{each fine voxel index $i$ in $^{t}\mathcal{C}$}
    \State \# Update vacancy probability of $i$
    \State $p(i) \leftarrow 0.5$
    \For{$k=t-n,...,t-1$}
        \If{$\exists i \in {^{k}\mathcal{Z}}$}
            \State update $p(i)$  as Eq. \ref{eq:dynamic_update}
        \EndIf
    \EndFor
    \State \# Segment dynamic points in $i$
    \If{$p(i) > 0.5$}
        \State $P^{d}_i \leftarrow P^{ng}_i$
    \EndIf
\EndFor
\State \# Sliding window for sequential dynamic removal
\State Delete $^{t-n}\mathcal{Z}$ in $\mathbb{Z}$
\State Add current measurement $^{t}\mathcal{Z}$ in $\mathbb{Z}$
\end{algorithmic}
\end{algorithm}

%% file: 4.5_DynaSTD.tex
\section{Dynamic-Aware loop closure}

This section introduces a dynamic-aware loop detection module. This module consists of two submodules: global scene description and hash table-based loop closure.

\subsection{Dynamic-aware scene description} \label{subsec:4_D}

Based on our online \ac{DOR} module in Sec. \ref{sec:1}, we integrate this module with STD \cite{yuan2023std}, which shares the same voxel representation. 
As STD extracts keypoints from boundary voxels where adjacent to plane voxels, a triangle descriptor $\Delta$ is formed by a combination of three neighboring keypoints as shown in Fig. \ref{fig:dyna_module2}.
$\Delta$ consists of six attributes: three side lengths of the descriptor, which are arranged in ascending order and dot products of projection normal vectors.
All $\Delta$ created from a single keyframe comprise the global descriptor $\mathbf{d}=\{\Delta_1,…,\Delta_l\}$ for the scene and each $\Delta$ are stored in local descriptor hash table $\mathbf{H}$. 

The combination of multiple local descriptors allows for matching even if only parts of two scenes overlap. This enables matching in scenarios with reverse loops or significant transition differences, particularly in the case of \ac{LiDAR} has a limited \ac{FoV}. It also facilitates matching across LiDAR modalities with different \ac{FoV}, as long as the same feature points and local triangle descriptors can be generated from local geometric characteristics.

However, dynamic objects that can be detected as keypoints act as outliers in encoding the consistent map structure. As shown in Fig. \ref{fig:dyna_module2}, (e) has false key points by pedestrians and produces a number of false links between keypoints. On the contrary, Fig. \ref{fig:dyna_module2} (f) demonstrates that false keypoints are rejected by adopting \ac{DOR}, and triangle descriptors from static regions are strengthened by connecting conservative keypoints. By eliminating dynamic objects, it is possible to generate a more discriminative global descriptor robust in high dynamic conditions.

\subsection{Loop closure based on DynaSTD} \label{subsec:5_A}

As all $\Delta$ is stored in $\mathbf{H}$, we adopt a hash function-based loop detection algorithm \cite{yuan2023std} where six attributes of $\Delta$ are utilized as a hash key. 
After selecting candidate loop pairs by voting manner and geometric verification of calculating plane overlap ratio, loop detection function $\mathcal{L}(\cdot)$ and relative pose aligner $\mathcal{S(\cdot)}$ are represented as follows:
\begin{equation}
\label{eq:std_detection}
\mathcal{L}:\left(\mathbf{d}_i, \mathbf{H}\right) \rightarrow\left(j, \mathbf{T}_{ij}\right), \quad \mathbf{T}_{ij}=\mathcal{S}\left(\mathbf{d}_i, \mathbf{d}_j\right),
\end{equation}
where descriptor $\mathbf{d}_i$ for query frame $i$ and $\mathbf{H}$ as an input, loop detector $\mathcal{L(\cdot)}$ finds a matched loop frame $j$ and relative pose $\mathbf{T}_{ij} \in \mathrm{SE}(3)$. 
As all vertices of matched $\Delta$ are paired by the uniqueness of side length, $\mathbf{T}_{ij}$ can be estimated by the point cloud registration problem with known data association. Relative pose estimation based on singular value decomposition (SVD) is as below:
\begin{equation}
\begin{aligned}
& \mathbf{A}=\sum_{i=1}^3\left(\mathbf{p}^i_{k}-\overline{\mathbf{p}}^i\right)\left(\mathbf{p}^j_{k}-\overline{\mathbf{p}}^j\right) \\
& {[\mathbf{U}, \mathbf{S}, \mathbf{V}]=\operatorname{SVD}(\mathbf{A})} \\
& \mathbf{R}_{ij}=\mathbf{V}\mathbf{U}^T \in \mathrm{SO}(3), \mathbf{t}_{ij}=-\mathbf{R}_{ij} \overline{\mathbf{p}}^i+\overline{\mathbf{p}}^j \in \mathbb{R}^3,
\end{aligned}
\end{equation}
which only utilizes triangle descriptors to estimate relative pose $\mathbf{T}_{ij} = [\mathbf{R}_{ij} \mid \mathbf{t}_{ij}]$ without scan matching. 
As conventional \ac{ICP} \cite{besl1992method}-based scan matching is not only time-consuming but also highly sensitive to the distribution of point cloud, the triangle descriptor-based pose estimator using local features enables effective alignment across multi-modal \ac{LiDAR} systems.

\subsection{Intra-session pose graph optimization} \label{subsec:4_C}
By dynamic aware scene description, keyframe map data $\mathbf{M}=\langle \mathbf{X}, \mathbf{S}, \mathbf{D},\mathbf{H} \rangle$ is processed where $\mathbf{X}=\{\mathbf{x}_k\}$ as $\mathrm{SE}(3)$ keyframe poses, $\mathbf{S}=\{\mathbf{s}_k\}$ as a set of static \ac{LiDAR} scans, $\mathbf{D}=\{\mathbf{d}_k\}$ as global descriptors and $k\in \mathbf{K}$ as keyframe index, respectively. Let multiple number of map $\mathbb{N}=\{1,...,\mathcal{N}\}$ and set of multiple map data as $\mathbb{M}=\{\mathbf{M}^1,...,\mathbf{M}^{\mathcal{N}}\}$.

As $\mathbf{X}$ from \ac{LiDAR} odometry contains its drift errors due to sensor measurement and processing noise, loop closing is essential to correct undesired effects.
Intra-session pose graph optimization of each $\mathcal{N}$ map is represented as:
\begin{equation}
\begin{aligned}
\hat{\mathbb{X}}=\underset{\mathbb{X}}{\operatorname{argmin}} & \sum_{m \in \mathbb{N}} \mathcal{F}(\mathbf{X}^m) ,
\end{aligned}
\end{equation}
with $\hat{\mathbb{X}}$ represents the set of optimized trajectories of multiple maps and single pose graph $\mathcal{F(\cdot)}$ is defined as:
\begin{equation}
\begin{aligned}
\mathcal{F}(\mathbf{X})= & \sum_{\langle t, t+1 \rangle \in \mathbf{K}}\left\|f\left(\mathbf{x}_t, \mathbf{x}_{t+1}\right)-\hat{\mathbf{z}}_{t, t+1}\right\|_{\sum_t}^2 \\
& +\sum_{\langle i, j \rangle \in \mathbb{L}^{intra}} \rho \left(\left\|f\left(\mathbf{x}_i, \mathbf{x}_j\right)-\mathcal{S}(\mathbf{d}_i,\mathbf{d}_j) \right\|_{\sum_{i, j}}^2 \right),
\end{aligned}
\end{equation}
where function $f:(\mathbf{x}_a,\mathbf{x}_b) \rightarrow (\mathbf{x}_b \ominus \mathbf{x}_a)$ as relative pose calculator, $\hat{\mathbf{z}}$ as estimated odometry, and $\mathbb{L}^{intra}$ as intra-session loop pairs by \ac{PR}. As $\mathcal{F}(\cdot)$ consists of the intra-session factor graph of each map, pose graphs are optimized independently. To handle incorrect loop constraints, robust kernel function~\cite{barron2019general} $\rho(\cdot)$ is utilized as:
\begin{equation}
\rho(e)=\xi^2 \log \left(1+\frac{e^2}{\xi^2}\right),
\end{equation}
where $\xi$ is the scale parameter and $e$ is the error residual of loop constraints.
Although $\hat{\mathbb{X}}$ demonstrates a reduction in drift error by pose graph optimization, the remaining drift may persist in scenarios where loops are absent, or loop pairs are too sparse to cover all trajectory error correction.
Also, each $\mathbf{X}$ is represented within its map coordinates, which is unscalable for a larger unified map. 

%% file: 5_MapMerge.tex
\section{Multiple Pose Graph Optimization}

Our multi-map merging module introduces a centralized map alignment method, including intra-session loop closure and multiple map alignment.    

\subsection{Multi-modal LiDAR map merging} \label{subsec:5_C}
Our multi-map merging module employs a centralized approach to optimize multiple pose graphs, accommodating both multi-session (to address temporal variances) and multi-robot (to handle spatial variances) scenarios.
Denote $\mathbf{M}^C$ as the central map and $\mathbf{M}^Q$ as the query map where the query map index $Q \in \mathbb{Q}=\mathbb{N}-\{{C}\}$. $\mathbf{M}^C$ is characterized by two conditions:
\begin{itemize}
\item Covers a wide area with a sufficient number of intra-session loop pairs.
\item The use of LiDAR with a wide \ac{FoV}, capable of generating a rich set of local descriptors.
\end{itemize}
Based on the above two conditions, the central map is considered sufficiently accurate and can serve as the target for aligning query maps.
Following the section \ref{subsec:4_C}, the query map data, including optimized robot pose $\mathbf{\hat{X}}^Q$, is shared with the central map server and inter-map loop constraints are detected to integrate multiple maps into a unified coordinate system.

Since each map is represented in its unique coordinate system before alignment, DynaSTD faces a discretization issue in inter-map loop detection. This problem arises because the keypoint extraction and identification of the global descriptor are processed in independent coordinate systems, stemming from its reliance on voxelization within each map's coordinate system. To alleviate this discrepancy, all $\Delta$ of $\mathbf{D}$ and $\mathbf{H}$ are refined as an ego-centric keyframe coordinate system for inter-map loop detection.

\begin{figure}[t]
	\centering
	\def\width{1.0\columnwidth}%
    {
    \includegraphics[clip, trim= 0cm 0cm 0cm 0cm, width=\width]{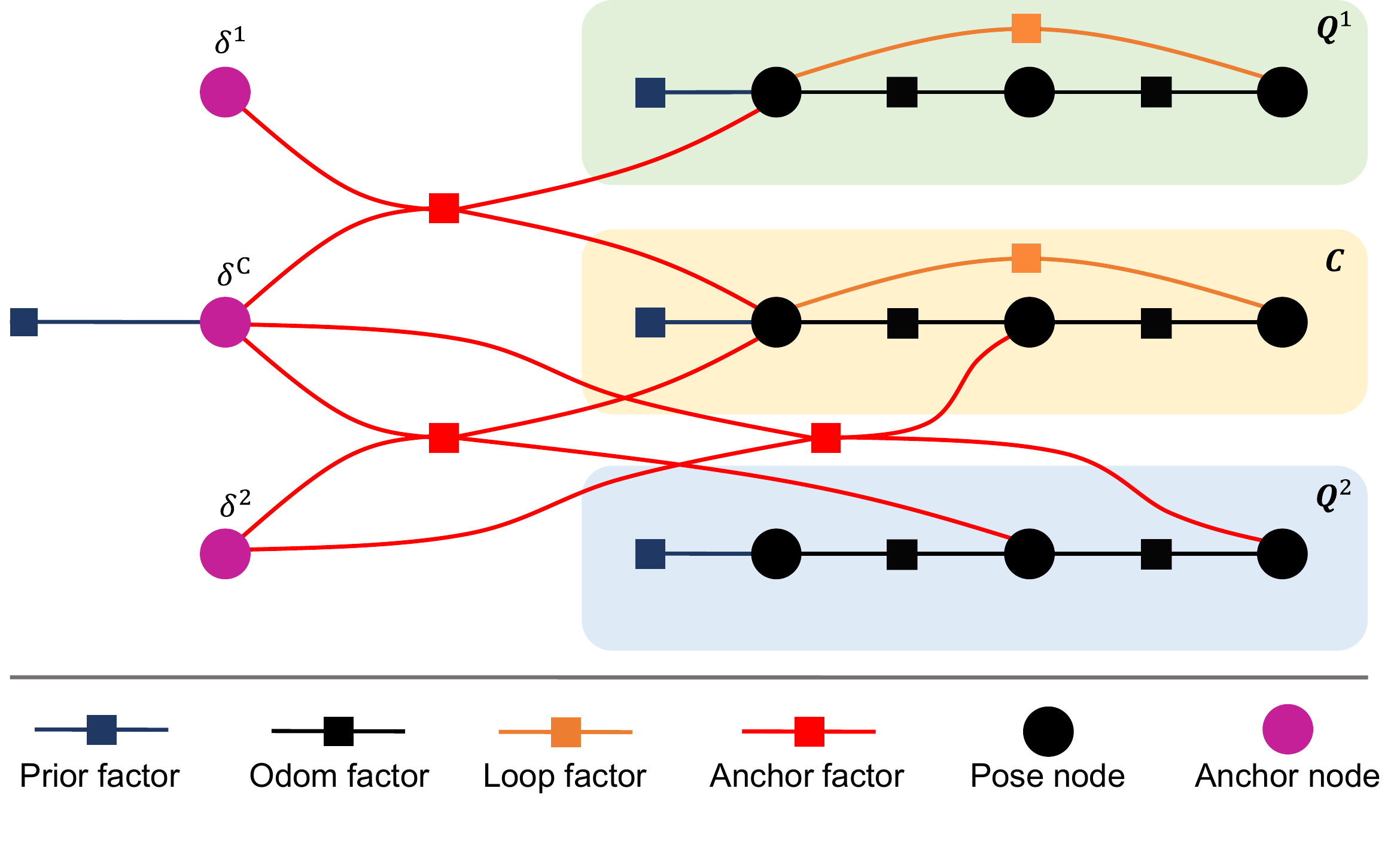}
	}
    \captionsetup{font=footnotesize}
 
    \caption{Illustration of anchor-node-based centralized multiple pose graph optimization. The anchor-node acts as an aligner from the query map to a central map coordinate and reduces drift error remaining where intra-session loop constraints are not present, similar to a map $Q^2$ }
	\label{fig:factor}
    \vspace{-0.45cm}
\end{figure}

As our dynamic-aware descriptor $\mathbf{D}$ contains the local feature of static structure, it is capable of searching inter-loop constraints in case of large differences in sensor modality, such as spinning type and solid-state types, which have little overlapping due to differences in \ac{FoV}. From this loop constraints robust to dynamics and sensor modality, anchor-node-based centralized multiple pose graph optimization is represented as:
\begin{equation}
\begin{aligned}
\mathbb{X}^*=\underset{\hat{\mathbb{X}}}{\operatorname{argmin}} \Big\{ 
\sum_{m \in \mathbb{N}} \mathcal{F}(\hat{\mathbf{X}}^m) + \sum_{Q\in \mathbb{Q}}\mathcal{A}(\hat{\mathbf{X}}^C,\hat{\mathbf{X}}^Q) \Big\}    ,
\end{aligned}
\end{equation}
where $\mathbb{X}^*$ is the pose set aligned with the unified coordinate of the central map. Anchor node-based factor graph $\mathcal{A}(\cdot)$ and anchor factor $\Phi(\cdot)$ are represented as:
\begin{equation}
\begin{aligned}
\mathcal{A}(\hat{\mathbf{X}}^C, \hat{\mathbf{X}}^Q)=
& \sum_{\langle i, j \rangle \in \mathbb{L}^{inter}} \left\| \Phi\left(\hat{\mathbf{x}}^C_{i}, \hat{\mathbf{x}}^Q_{j}, \delta^C, \delta^Q\right)- \hat{\mathbf{z}}_{i,j}
\right\|_{\sum_{i, j}}^2 \\
\end{aligned}
\end{equation}
\begin{equation}
\begin{aligned}
\Phi\left(\mathbf{x}^C_{i}, \mathbf{x}^Q_{j}, \delta^C, \delta^Q\right)=
\left(\left(\delta^C \oplus \mathbf{x}^C_{i}\right) \ominus\left(\delta^Q \oplus \mathbf{x}^Q_{j}\right)\right),
\end{aligned}
\end{equation}
where $\Phi(\cdot)$ constraints inter-map loop pairs with anchor nodes $\delta$  and poses of each map. 
SE(3) pose operators are denoted as $\oplus$ and $\ominus$.
As shown in Fig. \ref{fig:factor}, anchor factors represent the relative transformation between each map's origin coordinate systems to the centralized target system. By allocating a large covariance to query anchor nodes, all query anchor nodes are aligned to the central map to minimize measurement errors. 

Though our loop detection module can handle various types of \ac{LiDAR} sensors, it is still harder to find abundant loop pairs across modalities than in a homogeneous setting, as reported by \citet{jung2023helipr}. To achieve precise map merging, we implement a two-stage registration process, consisting of initial pose estimation and refinement registration steps. 
For the initial registration, we use Eq. (\ref{eq:std_detection}) to detect inter-loop pairing from the query map to the central map for map alignment. After a preliminary alignment, radius-based loop searching $\mathcal{R(\cdot)}$ is conducted to identify nearby loop constraints across all maps as follows:
\begin{equation}
\mathcal{R}:\left(\mathbf{M}^{t}, \mathbf{M}^s\right) \rightarrow\left(\mathbb{L}^R,\left\{\mathbf{T}_{i^{t} j^s}\right\}\right)
\end{equation}
\begin{equation}
\mathbb{L}^R=\left\{\langle i^t, j^s \rangle \mid \left\| \mathbf{x}^t_i \ominus \mathbf{x}^s_j  \right\|^2 < {r_{th}}^2 \right\},
\end{equation}
where $t$ and $s$ denote the target and source map indices, respectively, $r_{th}$ is the radius threshold, and $\mathbb{L}^R$ represents the total set of loop pairs from radius-based loop searching. 
For refined registration, G-ICP \cite{segal2009generalized} is adopted for point cloud registration. This two-phase map merging process ensures robust alignment, not just from query to central maps but also between query maps.

At the last stage, aligned maps are voxelized with respect to a central map coordinate, and each voxel constructs its own kd-tree for local point cloud. This structure enables efficient detection of previously unmapped or newly added regions by performing nearest neighbor (NN) searches within individual voxels. The time complexity of these NN searches is reduced to $O(\log N_v)$, where $N_v$ is the number of points in a voxel. By limiting kd-tree construction and queries to relevant voxels, Uni-Mapper substantially reduces computational cost and supports scalable, dynamic map merging.

%% file: 6_EvaluationSetup.tex
\section{Evaluation Setup}

In this section, we present the datasets used for experiments and the evaluation criteria on dynamic object removal, place recognition, and multiple map merging. Also, brief reviews of comparision \ac{SOTA} methodologies are enumerated. All experiments were conducted on an Intel Core i7-12700KF processor with 64GB of RAM. Our framework is fully implemented in C++ with ROS Noetic version on Ubuntu 20.04.

\subsection{Datasets}

To evaluate the performance of algorithms, we utilized nine sequences from two public datasets (\textit{HeLiPR} and \textit{Semantic KITTI}) and our custom dataset (\textit{INHA}), all of which feature various types of \ac{LiDAR} modalities across multi-session and multi-robot scenarios. Each dataset captures dynamic environments populated with moving vehicles and pedestrians. The specific characteristics of each data sequence will be detailed in subsequent subsections. Additionally, a comprehensive summary of each data sequence is provided in Table \ref{tab:dataset}.

\subsubsection{HeLiPR Dataset}
The \textit{HeLiPR} dataset \cite{jung2023helipr} encompasses a diverse array of range sensors, including a 128-ray scanning \ac{LiDAR} (Ouster OS2-128), a 16-ray scanning \ac{LiDAR} (Velodyne VLP-16), a 64-ray solid-state \ac{LiDAR} (Aeva AeriesII), and a 6-ray non-repetitive \ac{LiDAR} (Livox Avia). 
Each \ac{LiDAR} sensor exhibits a wide range of \ac{FoV}, spanning from $70^{\circ}$ to $360^{\circ}$ horizontally and from $19.2^{\circ}$ to $77^{\circ}$ vertically. 
Furthermore, the dataset provides ground-truth poses for each \ac{LiDAR} sensor, derived from a high-precision inertial navigation system (INS). 
Among the multi-session data sequences from three locations, we use the \texttt{Town} sequence, showcasing diverse urban environments with wide and narrow roads, as well as dynamic elements such as buses, vehicles, and pedestrians. 
As  \texttt{Town} comprises three distinct sessions with four different LiDAR sensors, we have a total of twelve distinct data sequences. For convenience, each data sequence is referred to using a naming convention that combines the sensor name with the sequence number. For example, the Aeva LiDAR data from the first session will be designated as \texttt{Aeva1}. This naming convention will be consistently applied throughout the discussion in the subsequent sections of the paper.

\begin{table*}[t]
\centering
\caption{Dataset for multi-session and multi-map merging}
\label{tab:dataset}
\resizebox{\textwidth}{!}{%
\begin{tabular}{c|ccc|cccc}
\hline\hline
\textbf{Dataset} &
  \multicolumn{3}{c|}{\textit{HeLiPR}} &
  \multicolumn{4}{c}{\textit{INHA}} \\ \hline
\textbf{Sequence} &
  \multicolumn{3}{c|}{\texttt{TOWN}} &
  \multicolumn{1}{c|}{\texttt{WHEEL}} &
  \multicolumn{1}{c|}{\texttt{DOG}} &
  \multicolumn{1}{c|}{\texttt{HAND1}} &
   \texttt{HAND2} \\ \hline\hline

\multirow{2}{*}[2em]{\textbf{Platform}} &
  \multicolumn{3}{c|}{\includegraphics[height=1.5cm]{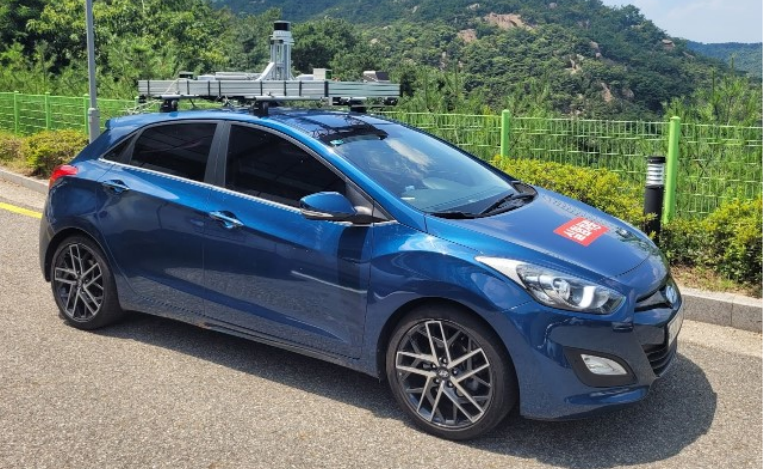}} &
  \multicolumn{1}{c|}{\includegraphics[height=1.5cm]{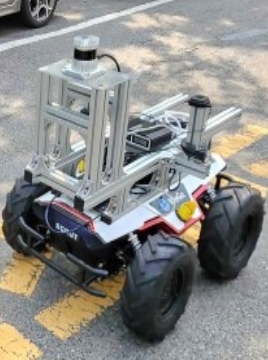}} &
  \multicolumn{1}{c|}{\includegraphics[height=1.5cm]{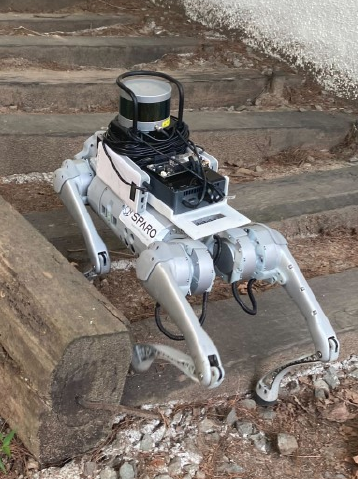}} &
  \multicolumn{1}{c|}{\includegraphics[height=1.5cm]{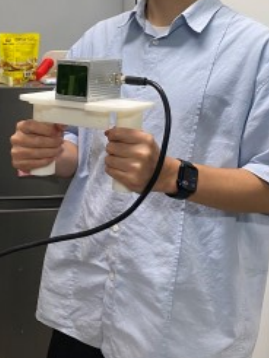}} &
  \includegraphics[height=1.5cm]{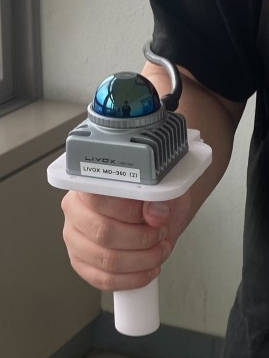}
   \\ 
 &
  \multicolumn{3}{c|}{Vehicle} &
  \multicolumn{1}{c|}{AgileX-ScoutV2} &
  \multicolumn{1}{c|}{Unitree-Go1} &
  \multicolumn{1}{c|}{Handheld} &
  Handheld \\ \hline
\multirow{2}{*}[2em]{\textbf{LiDAR}} &
  \multicolumn{1}{c|}{\includegraphics[width=1.5cm]{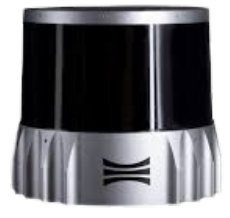}} &
  \multicolumn{1}{c|}{\includegraphics[width=1.5cm]{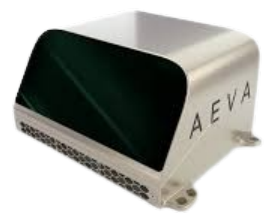}} &
  \multicolumn{1}{c|}{\includegraphics[width=1.5cm]{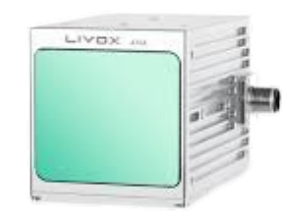}} &
  \multicolumn{1}{c|}{\includegraphics[width=1.5cm]{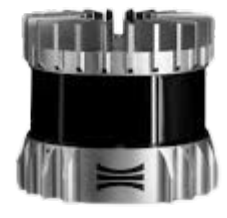}} &
  \multicolumn{1}{c|}{\includegraphics[width=1.5cm]{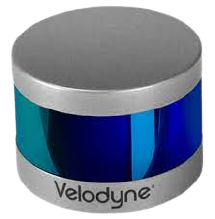}} &
  \multicolumn{1}{c|}{\includegraphics[width=1.5cm]{fig/lidar/avia.pdf}} &
  \includegraphics[width=1.5cm]{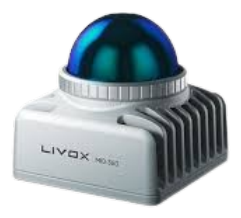}
   \\
 &
  \multicolumn{1}{c|}{Ouster OS2-128} &
  \multicolumn{1}{c|}{Aeva Aeries II} &
  Livox Avia &
  \multicolumn{1}{c|}{Ouster OS1-64} &
  \multicolumn{1}{c|}{Velodyne VLP-16} &
  \multicolumn{1}{c|}{Livox Avia} &
  Livox Mid-360 \\ \hline
\textbf{FoV(H$\times$V)} &
  \multicolumn{1}{c|}{$360^{\circ}$ $\times$ $22.5^{\circ}$} &
  \multicolumn{1}{c|}{$120^{\circ}$ $\times$ $19.2^{\circ}$} &
  $70^{\circ}$ $\times$ $77^{\circ}$ &
  \multicolumn{1}{c|}{$360^{\circ}$ $\times$ $45^{\circ}$} &
  \multicolumn{1}{c|}{$360^{\circ}$ $\times$ $30^{\circ}$} &
  \multicolumn{1}{c|}{$70^{\circ}$ $\times$ $77^{\circ}$} &
  $360^{\circ}$ $\times$ $59^{\circ}$ \\ \hline
\textbf{Type} &
  \multicolumn{1}{c|}{spinning / 128ch} &
  \multicolumn{1}{c|}{solid state / 64ch} &
   solid state / 6ch&
  \multicolumn{1}{c|}{spinning / 64ch} &
  \multicolumn{1}{c|}{spinning / 16ch} &
  \multicolumn{1}{c|}{solid state / 6ch} &
   solid state / 40ch\\ \hline
\textbf{Dyna Objs} &
  \multicolumn{1}{c|}{$\bigstar$$\bigstar$$\bigstar$} &
  \multicolumn{1}{c|}{$\bigstar$$\bigstar$$\bigstar$} &
   $\bigstar$$\bigstar$$\bigstar$&
  \multicolumn{1}{c|}{$\bigstar$$\bigstar$} &
  \multicolumn{1}{c|}{$\bigstar$$\bigstar$$\bigstar$} &
  \multicolumn{1}{c|}{$\bigstar$} &
   $\bigstar$\\ \hline
\end{tabular}%
}
\end{table*}

\subsubsection{INHA Dataset}
Our custom dataset named \textit{INHA} is specifically designed to validate the effectiveness of our proposed methodology for a multi-robot and multi-modal \ac{LiDAR} {SLAM} framework. 
As illustrated in Table \ref{tab:dataset}, our multi-agent robot team consists of an Agile-X Scout-V2, Unitree Go-1, and custom hand-held mapping systems. 
Each mapping platform is equipped with its own \ac{LiDAR} sensor: a 64-ray scanning \ac{LiDAR} (Ouster OS1-64), a 16-ray scanning \ac{LiDAR} (Velodyne VLP-16), a 6-ray non-repetitive \ac{LiDAR} (Livox Avia) and 40-ray non-repetitive \ac{LiDAR} (Livox Mid-360). Each sequence is named by the robot platform as \texttt{WHEEL}, \texttt{DOG}, \texttt{HAND1}, and \texttt{HAND2}, respectively.
\texttt{WHEEL} is designated as a central sequence and others have overlapping areas with it. 
Each sequence has unique environmental features such as steep stairs, indoor-to-outdoor, and multi-floor, necessitating the use of appropriately suited robots for mapping.  
Given the practical challenges of mounting an INS system on robot platforms and the unavailability of GPS signals in certain areas, we generate ground truth trajectories using interactive SLAM \cite{koide2020interactive}. 

\subsubsection{Semantic KITTI Dataset}
As both multi-modal \ac{LiDAR} datasets (\textit{HeLiPR}, \textit{INHA}) lack of point-wise semantic information, we utilize the \textit{Semantic KITTI} dataset \cite{behley2019semantickitti} for evaluation of dynamic object removal. 
\textit{Semantic KITTI} utilizes a 64-ray scanning \ac{LiDAR} (Velodyne HDL-64E) with a vertical \ac{FoV} of $30^{\circ}$ and $360^{\circ}$ in horizontal.
As \textit{Semantic KITTI} provides point-wise ground truth labels for both dynamic and non-moving objects, we evaluate the quantitative results of our dynamic removal module in sequence \texttt{00}.

\subsection{Evaluation Metrics}

\subsubsection{Static mapping}
To evaluate the performance of dynamic object removal and static mapping, we utilize the open-source dynamic map benchmark \cite{zhang2023dynamic}. The evaluation metrics used in~\cite{zhang2023dynamic} are defined as follows:
\begin{equation}
\begin{aligned}
& \mathrm{SA}=\frac{N_{\text {True static }}}{N_{\text {Total static }}},
\, \mathrm{DA}=\frac{N_{\text {True dynamic }}}{N_{\text {Total dynamic }}},
\end{aligned}
\end{equation}
where $N$ denotes the number of point cloud. 
SA assesses the preservation of static areas, while DA gauges the accuracy of estimation for dynamic object removal. The geometric mean of SA and DA, represented by ${AA} = \sqrt{SA \times DA}$, indicates the overall performance of dynamic object removal and static map conservation. Given the challenge of fairly comparing all state-of-the-art algorithms due to individual voxelization approaches, we standardize the voxel leaf size to 0.1 for all models to ensure a uniform basis for evaluation.

\subsubsection{Place recognition}
To assess the effectiveness of the loop detection and underscore the importance of dynamic-aware and multi-modal \ac{LiDAR} scene descriptions, we employ the precision-recall curve. Precision (Pr), Recall (Re), and F1 score (F1) are defined \cite{yin2023survey} as:
\begin{equation}
\text {Pr}=\frac{\mathrm{TP}}{\mathrm{TP}+\mathrm{FP}},
\, \text { Re}=\frac{\mathrm{TP}}{\mathrm{TP}+\mathrm{FN}},
\, \text { F1}=2\times\frac{\mathrm{Pr} \times \mathrm{Re}}{\mathrm{Pr}+\mathrm{Re}}
\end{equation}
where TP represents true positives, FP denotes false positives, and FN stands for false negatives. 
The F1 score is the harmonic mean of precision and recall, which implies the accuracy of the model.
Considering that the judgement of the correct loop matching is denoted as $\text{GT}=\text{TP}+\text{FN}$, influenced by the distance threshold $d_{th}$ between two paired poses. We establish $d_{th}$ at 35m for the \textit{HeLiPR} and 20m for \textit{INHA} datasets.

\subsubsection{Multiple map merging}
To evaluate the effectiveness of pose graph optimization in both intra-session and multiple map scenarios, we adopt the Absolute Trajectory Error (ATE) evaluation metric. 
ATE is adopted to measure the ability of multiple map merging by calculating the relative pose transformation from the estimated pose to the ground truth pose on global coordinates as:
\begin{equation}
e_{\text {abs}}(\mathbf{X})=\frac{1}{N} \sum_{i \in N}\left\|\left(\mathbf{x}^*_i \ominus \mathbf{x}^{\star}_i\right) \right\|_2,
\end{equation}
where $\mathbf{x}^{\star}$ is defined as the ground truth SE(3) pose and $\mathbf{x}^{*}$ as the estimated pose.
We utilize open-source benchmark tools~\cite{grupp2017evo} to evaluate the above metric for multiple sequences.

To assess the quality of the merged map, we employ Accuracy (AC) \cite{seitz2006comparison}, Chamfer distance (CD) \cite{wu2021density}, and Mean Map Entropy (MME) \cite{droeschel2014local}.
Let the aligned point cloud maps be denoted as $P_1=\{x\}$, $P_2=\{y\}$. Inlier points set $P_1'$ and $P_2'$ is defined as the subsets of $P_1$ and $P_2$ containing points whose nearest neighbors in the other cloud are within a threshold distance $\tau$.
Since the maps are partially overlapping and there may be trajectory errors in the aligned poses, we set a distance threshold $\tau$ of 5m to account for uncertainties in the registration. 

The accuracy is defined as the mean euclidean distance of the inlier point pairs as $\text{AC} = \sqrt{ \frac{1}{|I|} \sum_{(x,y) \in \mathcal{I}} \|x - y\|^2 }$ where \( \mathcal{I} = \{(x, y) | \|x - y\| < \tau \} \)
and CD, representing the quality of registration between the two point cloud maps, is calculated as:
\begin{equation}
\text{CD} = \frac{1}{|P_1'|} \sum_{x \in P_1'} \min_{y \in P_2} \|x - y\|_2^2 + \frac{1}{|P_2'|} \sum_{y \in P_2'} \min_{x \in P_1} \|x - y\|_2^2.
\end{equation}
The MME metric, which reflects the point-wise local consistency of aligned maps, is calculated as:
\begin{equation}
\text{MME} = \frac{1}{|P_1|+|P_2|} \sum_{z \in P_1 \cup P_2 } \left[ \frac{1}{2} \ln \left( 2\pi e \Sigma (z) \right) \right],
\end{equation}
where $\Sigma (z)$ represents the covariance of the neighbor points in the map. We use the same threshold $\tau$ to determine the set of neighbor points.

\subsection{Comparing Methods}

\subsubsection{Dynamic Object Removal}
Since dynamic object removal algorithms can be broadly categorized by data representation, we select representative methods for each category: OctoMap \cite{hornung2013octomap}, Dynablox \cite{schmid2023dynablox}, and DufoMap \cite{duberg2024dufomap} for occupancy-based approaches, Removert \cite{kim2020remove}, ERASOR~\cite{lim2021erasor}, and BeautyMap \cite{jia2024beautymap} for discrepancy-based methods. 
For \textit{Semantic KITTI} dataset, all configurations of each comparison method follow the setting introduced in \cite{zhang2023dynamic}. For other datasets such as \textit{HeLiPR} or \textit{INHA}, we set the \ac{FoV} parameter and sensor height for each LiDAR setting and mobility platform. 

\subsubsection{Place Recognition}
We compare \ac{SOTA} methodologies that could leverage open-source implementation for the experiments, namely M2DP \cite{he2016m2dp}, SC \cite{kim2018scan}, STD \cite{yuan2023std} for algorithm-based methods and OT \cite{ma2022overlaptransformer}, LoGG3D-Net \cite{vidanapathirana2022logg3d} for learning-based methods with our proposed method. For the evaluation of SC, we utilized a C++ implementation from an author with the same configuration except for the search ratio parameter. As this parameter restricts the searching ratio, which is for fast loop searching but vulnerable to rotational variances, we set this parameter as 1 to search all possible rotation directions. 
For OT and LOGG3D-Net, we use the model weights released by the authors, which were pretrained for evaluation purposes.
As M2DP, SC, and LoGG3D-Net are based on single-shot \ac{LiDAR} scan, we utilize downsampled single frame point clouds from FAST-LIO2 \cite{xu2021fast} as input data, except for non-repetitive scanning LiDARs such as Livox's Avia or Mid-360. For these non-repetitive LiDAR types, we compensate for sparsity by utilizing submaps accumulated over 10 frames. Both STD and our method accumulate downsampled point clouds for all types of \ac{LiDAR} point clouds as keyframe data. We sample 5m for target sequences and 10m for query sequences in the \textit{HeLiPR} dataset. For \textit{INHA} datasets, all keyframe point cloud are utilized without any distance-based sampling.

\subsubsection{Multi-map alignment}
For the evaluation of multi-map alignment, we utilize LT-mapper \cite{kim2022lt}, which handles dynamic object removal and multi-session map alignment simultaneously, 
and LTA-OM \cite{zou2024lta} as a multi-session localization and map stitching framework.
All evaluations follow the code and the parameter settings provided in open-source. 
As our method utilizes G-ICP \cite{segal2009generalized} for a refined registration process, we re-implement the registration method of LT-mapper, replacing ICP with G-ICP, for a fair comparison.

%% file: 7_EvaluationResults.tex
\section{Evaluation Results}
\label{sec:7}

\subsection{Dynamic Object Removal}
To effectively use dynamic object removal as a pre-filtering module for place recognition in dynamic environments, both key conditions must be required: preservation of static elements and computational efficiency.

\begin{table*}[t!]
\centering
\caption{Quantitative evaluation of dynamic object removal in point cloud maps.}
\label{tab:removal}
\resizebox{\textwidth}{!}{%
\begin{tabular}{c|cccc|cccc|cccc}
\hline \hline
              & \multicolumn{4}{c|}{\textit{KITTI} \texttt{00}}                 & \multicolumn{4}{c|}{\textit{INHA} \texttt{WHEEL}}             & \multicolumn{4}{c}{\textit{HeLiPR} \texttt{Avia1}}                \\ \hline\hline
Method        & SA$\uparrow$ & DA$\uparrow$  & AA$\uparrow$  & Runtime{[}ms{]} & SA$\uparrow$ & DA$\uparrow$  & AA$\uparrow$  & Runtime{[}ms{]} & SA$\uparrow$      & DA$\uparrow$      & AA$\uparrow$      & Runtime{[}ms{]} \\ \hline
Removert      & 99.44 & 41.53 & 64.26 & 222                                    & 98.66 & 81.93 & 89.91 & 65                                     & 98.27 & 53.01 & 72.17 & \textbf{33}                            \\
ERASOR        & 84.07 & 97.80 & 90.67 & 132                                    & 94.82 & 83.96 & 89.22 & 72                                     & 99.47 & 18.66 & 43.08 & \bl{\textbf{25}}                            \\
BeautyMap     & 96.95 & 98.33 & \rl{\textbf{97.64}} & \bl{\textbf{36}}                            & 87.57 & 95.38 & \rl{\textbf{91.39}} & \bl{\textbf{29}}                            &   N/A   &   N/A   &   N/A   & N/A                             \\   
\hline
Octomap       & 76.78 & 99.46 & 87.39 & 1243                                   & 89.74 & 87.41 & 88.57 & 1658                                   & 88.58 & 90.47 & \rl{\textbf{89.52}} & 2598                 \\
DynaBlox      & 96.76 & 90.68 & 93.67 & 254                                    & 99.50 & 54.95 & 73.94 & 308                                    & 98.94 & 20.61 & 45.15 & 396                           \\
DUFOMap       & 90.63 & 98.99 & \bl{\textbf{94.71}} & \textbf{56}                         & 93.84 & 87.48 & \textbf{90.61} & \textbf{47}                            & 98.41 & 80.24 & \bl{\textbf{88.86}} & 57                \\
Ours          & 98.11 & 89.99 & \textbf{93.96} & \rl{\textbf{10}$^*$}               & 99.33 & 83.57 & \bl{\textbf{91.11}} & \rl{\textbf{14}$^*$}            & 98.82 & 69.81 & \textbf{83.06} & \rl{\textbf{11}$^*$}      \\ 
\hline
\end{tabular}%
}
\begin{minipage}{18.5cm}
\vspace{0.1cm}

The best, second best, and third best results are shown in \rl{\textbf{red}}, \bl{\textbf{blue}}, and {\textbf{black}} bold, respectively.
* means runtime is the sum of dynamic object removal and scene description.
\end{minipage}
\vspace{-0.2cm}

\end{table*}

\begin{figure*}[h!]
    \centering
	\def\width{1.9\columnwidth}%
	\includegraphics[width=\width, trim=0cm 0cm 0cm 0cm, clip]{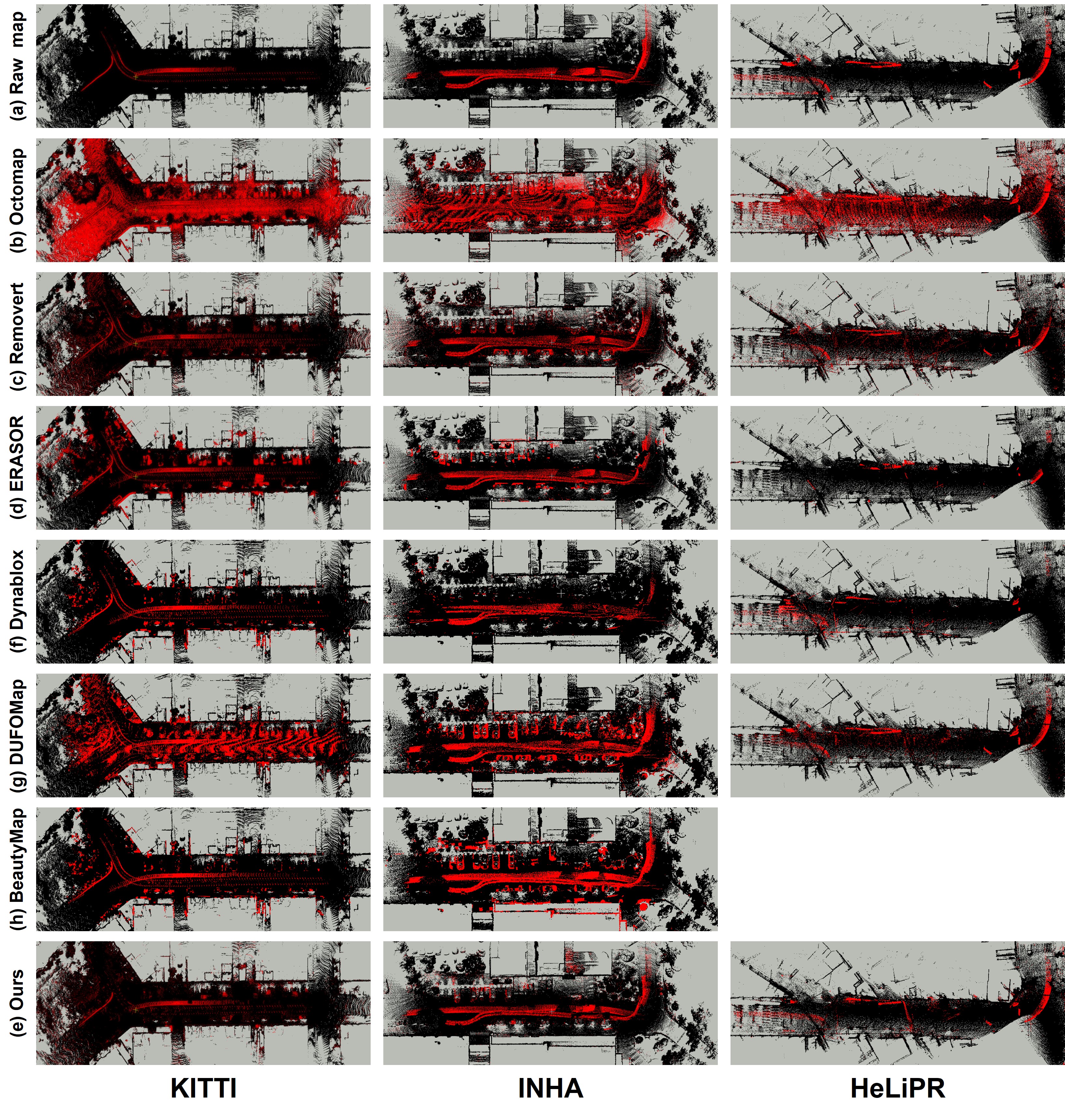}
    \vspace{-0.2cm}
 \captionsetup{font=footnotesize}
	\caption{Qualitative comparison of \ac{DOR} performance between proposed method and existing methods on sequence \texttt{00} of \textit{SemanticKITTI}, \texttt{Aeva1} of \textit{HeLiPR}, and \texttt{WHEEL} of \textit{INHA}. Red points indicate classified dynamic objects, and the black points as static objects.} 
    \label{fig:result}
\end{figure*}
Compared with map-based offline dynamic removal methods (Removert, ERASOR, BeautyMap) and online methods (Octomap, DynaBlox, DUFOMap), our dynamic removal module showed competitive performance in dynamic filtering while achieving the lowest computation time, which highlights its suitability for real-time robot operation. 
Its effectiveness is demonstrated both quantitatively and qualitatively on various datasets. 
Table \ref{tab:removal} and Fig. \ref{fig:result} present the performance and computational efficiency of dynamic object removal algorithms on sequence \texttt{00} (4390 - 4530) of \textit{SemanticKITTI}, \texttt{Avia1} (keyframe 295 - 320) of \textit{HeLiPR}, and \texttt{WHEEL} (keyframe 300 - 350) of \textit{INHA}. 
For \textit{HeLiPR} and \textit{INHA}, we manually labeled dynamic points using CloudCompare \cite{cc2023cc}.


As shown in Table \ref{tab:removal}, BeautyMap \cite{jia2024beautymap} achieves the highest AA score, demonstrating strong performance in dynamic object removal via binary encoding and matrix comparison. However, its reliance on a square matrix format causes failures with limited-\ac{FoV} LiDARs, such as in the \textit{HeLiPR} \texttt{Avia1} sequence, where much of the matrix is empty. Furthermore, its dependence on a pre-built map limits its ability to filter dynamic points in real time during robot operation.

While Removert~\cite{kim2020remove} shows a high SA score, which implies the performance of static points preservation, it relatively underperforms when removing dynamic points except for the \textit{INHA} dataset (DA 41 - 53\%). 
ERASOR \cite{lim2021erasor} shows satisfactory performance on \textit{KITTI} and \textit{INHA} datasets. However, in the case of \textit{HeLiPR}, where numerous electric wires are present above the road, segmenting dynamic bins becomes more challenging because the height discrepancy between the map and the scan is not affected by dynamic objects. 

For online removal methods that operate without a pre-built map, Octomap \cite{hornung2013octomap} exhibits the best performance in \textit{HeLiPR} sequence. However, as shown in Fig. \ref{fig:result} (b), the incidence angle ambiguity problem misclassifies large areas of the ground and static regions as dynamic points and removes them, resulting in a lower SA score. Furthermore, the heavy computation involved in point-wise ray tracing makes it the most time-consuming process and unsuitable for real-time operations.
Although DynaBlox shows better computational efficiency compared to Octomap, it still struggles to operate in real-time with dense point clouds. 

DUFOMap shows improved computational performance due to the UFOMap structure and effectively removes dynamic points regardless of LiDAR types. However, as shown in Fig. 8 (e), it removes static objects such as cars and trees, resulting in a relatively low SA score, which negatively impacts 3D scene description, particularly when using local features like STD.
Additionally, DUFOMap's dependency on UFOMap makes it challenging to integrate with other data structures of scene descriptors, requiring additional memory to perform place recognition simultaneously.

Our dynamic removal module, which shares the voxel structure of the scene descriptor, demonstrates superior computational efficiency while performing dynamic removal and scene description. 
Among online methods, our module achieves an SA score above 98 and the highest AA score on \textit{SemanticKITTI}, second on \textit{INHA}, and third on \textit{HeLiPR}, demonstrating well-balanced performance in both dynamic object removal and static map preservation. 

Our method's sliding window operation enables the correction of erroneous judgments when they are correctly identified in a subsequent frame, thereby reducing the possibility of mistakenly removing static areas. This emphasizes the efficiency of our coarse-to-fine free space hash map and dynamic segmentation algorithms, which share the same voxel representation with scene description.
Unlike other \ac{SOTA} methods, which need a prior map for dynamic object removal, our method demonstrates applicability even without precise prior pose or map information.

\begin{figure}[t]
    \centering
	\def\width{1\columnwidth}%
	\includegraphics[width=\width, trim=0 0 0 0, clip]{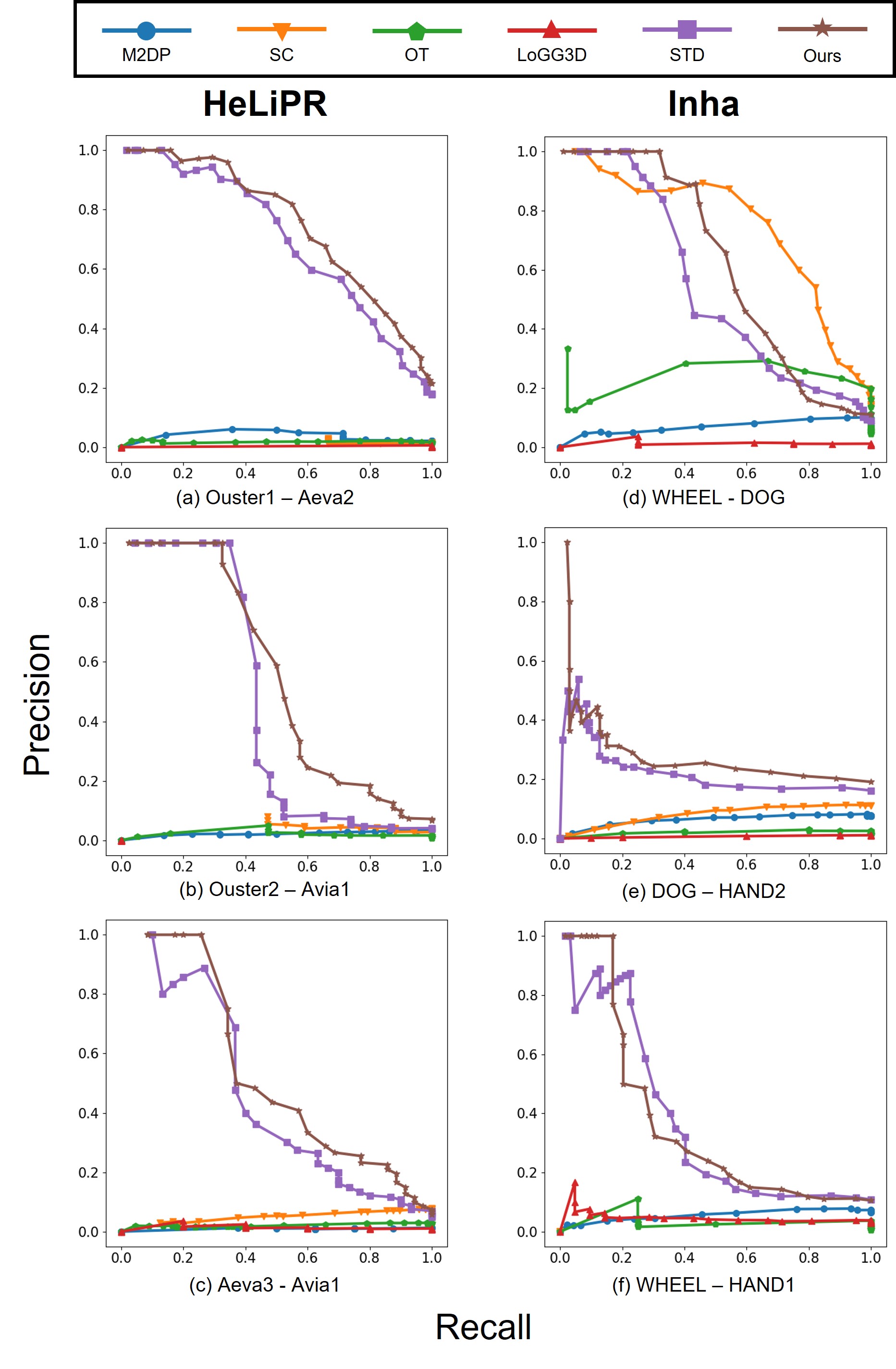}
        \captionsetup{font=footnotesize}
	\caption{Precision-Recall curve on \textit{HeLiPR} and \textit{INHA} datasets.}
    \label{fig:pr_curve}
\end{figure}


\subsection{Place recognition on multi sequences}
\label{subsec:7-B}

\subsubsection{Precision Recall Evaluation}
We first evaluated the loop detection performance using a precision-recall curve to verify the necessity of a dynamic-aware and LiDAR-modal-agnostic description. The precision-recall curves across multi-modal LiDARs are plotted in Fig. \ref{fig:pr_curve}. As can be seen, our method showed superior performance in multi-modal LiDAR settings and even in dynamic environments. 

Algorithm-based global descriptors such as M2DP and SC encode the overall scene, which is vulnerable to different \ac{FoV} in the case of loop detection between spinning and solid-state \ac{LiDAR}. 
In particular, SC shows the most significant performance gap depending on the LiDAR's \ac{FoV}. As shown in Fig. \ref{fig:pr_curve} (d), SC shows stable performance among spinning types (OS1-64 and VLP-16), both of which have an omnidirectional \ac{FoV}. However, the matching performance declined remarkably between spinning types (Ouster/Velodyne) and solid-state types (Aeva/Livox), which resulted in considerable differences in \ac{FoV} and scanning pattern. This is because SC generates 2D descriptors by encoding the highest z-value in a bin-wise manner. Differences in \ac{FoV} lead to a loss in the descriptor, and the ringkey of SC, which condenses the descriptor to a single vector, is distorted, resulting in a loss of the descriptor's discriminative power.

\begin{figure*}[t]
	\centering
	\def\width{0.97\textwidth}%
    {%
		\includegraphics[clip, trim= 0 5 30 0, width=\width]{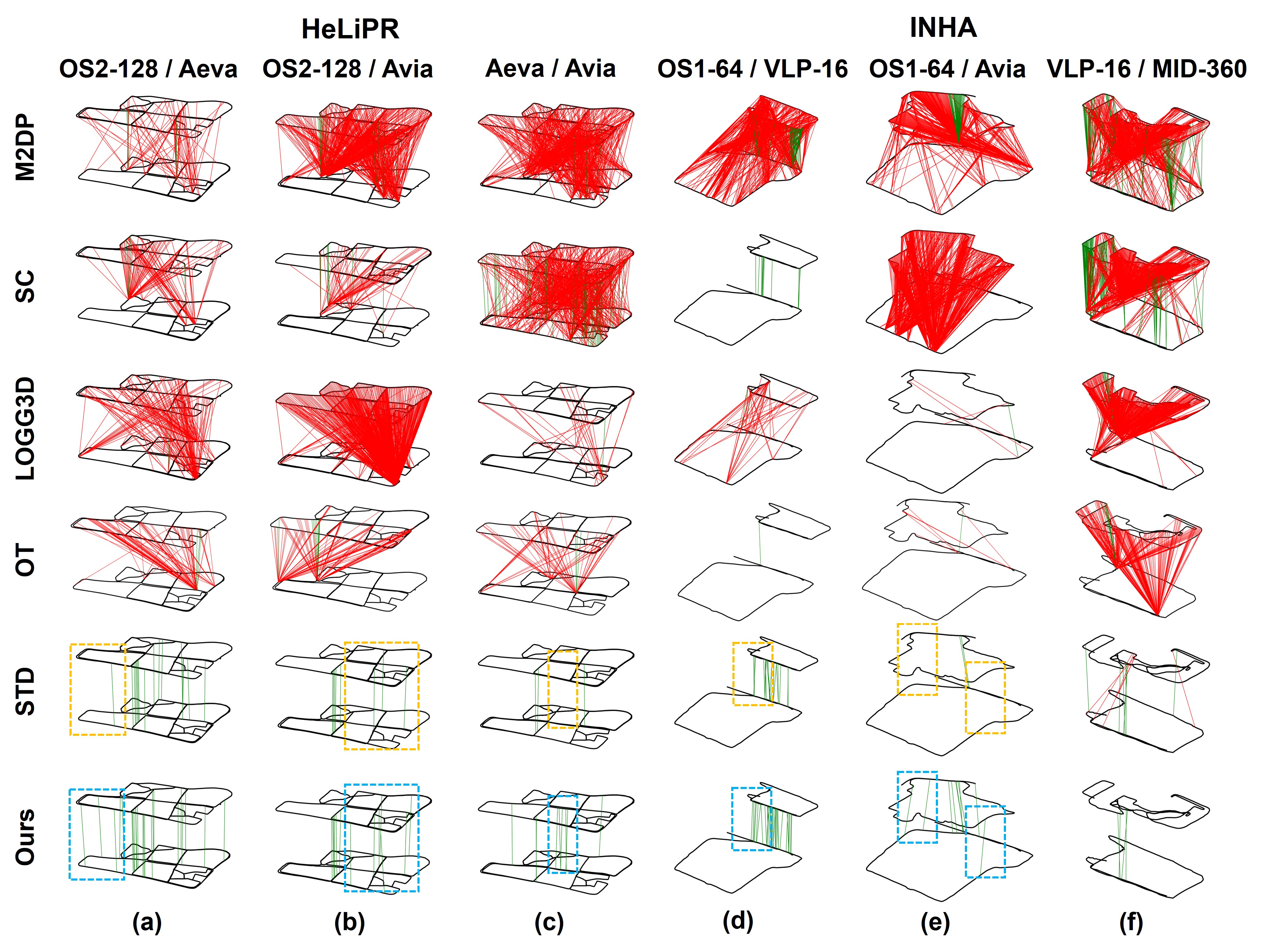}
	}
    \captionsetup{font=footnotesize}
    \caption{
    The place recognition results for multi-modal datasets under maximum precision. In each pair, the lower and upper trajectories represent the central and query maps, respectively. Sequences (a), (b), and (c) are pairs from the \textit{HeLiPR} dataset, while (d), (e), and (f) are from the \textit{INHA} dataset. Loop detection is performed from the query to the central map, with the green line indicating TP and the red line indicating FP.
    }
	\label{fig:main_result}
   \vspace{-0.5cm}
\end{figure*}

OT also shows significant dependency on LiDAR attributes such as the type of scanning or the number of channels, because it projects point cloud onto a range image. In contrast, LoGG3D-Net perceives LiDAR scans as collections of 3D point clouds irrespective of LiDAR types. However, LoGG3D-Net also demonstrates an inability to overcome differences in point cloud distribution due to different \ac{FoV} and scanning patterns. As \citet{yin2023survey} pointed out, maintaining consistent performance across different sensor configurations and in previously unencountered locations remains a challenge that learning-based methodologies ought to overcome.


STD and our method exhibit robust performance across different LiDAR modalities. 
This is because the global descriptor is formed by aggregating local triangle descriptors, allowing loop detection regardless of point cloud modality, as long as similar local features can be extracted. 
However, since these methods rely on local descriptors, our approach (DynaSTD) achieves better performance in dynamic environments by removing inconsistent local features.
As illustrated in Fig. \ref{fig:pr_curve}, our method maintains high discriminability under dynamic conditions, while still performing comparably in scenes with fewer dynamic objects (Fig. \ref{fig:pr_curve}(e)).


Overall, STD and our method outperform results reported in \textit{HeLiPR} \cite{jung2023helipr}, mainly due to the different loop-pair distance threshold $d_{th}$.  
While \citet{jung2023helipr} set $d_{th}$ to 7.5m, we adopt 35m, as it better reflects the characteristics of the dataset, including wide roads and reverse loops from a multi-session perspective.
Especially, STD and our method, suitable for narrow-\ac{FoV} solid-state \ac{LiDAR}, are particularly effective at detecting loops with large spatial gaps, as evidenced by higher precision under our threshold setting.

\subsubsection{Max precision loop matching}

\begin{table}[t]
\centering
\caption{{Matched loop pair in max precision}}
\label{tab:match_pair}
\resizebox{\columnwidth}{!}{%
\begin{tabular}{c||c|c|c|c|c}
\hline
Seq. & Method & Loops & \# of TP & Precision & \multicolumn{1}{c}{F1 Score} \\ \hline\hline
\multirow{6}{*}{\begin{tabular}[c]{@{}c@{}}\textit{HeLiPR}\\ (\texttt{Ouster1-Aeva2})\end{tabular}} 
 & M2DP & 76 & 5 & 6.57 & 0.100 \\
 & SC & 108 & 4 & 3.70 & 0.051 \\
 & OT & 93  & 3  & 3.23  & 0.049  \\
 & LoGG3D & 165  & 1  & 0.61  & 0.012  \\
 & STD & 16 & \underline{16} & \textbf{100} & \underline{0.625} \\
 & Ours & 23 & \textbf{23} & \textbf{100} & \textbf{0.660} \\ \hline\hline
\multirow{6}{*}{\begin{tabular}[c]{@{}c@{}}\textit{HeLiPR}\\ (\texttt{Ouster2-Avia1})\end{tabular}} 
 & M2DP & 629 & 22 & 3.49 & 0.068 \\
 & SC & 80 & 7 & 8.75 & 0.129 \\
 & OT & 164  & 9 & 5.49 & 0.098 \\
 & LoGG3D & 631  & 0 & 0 & 0.000 \\
 & STD & 8 & \underline{8} & \textbf{100} & \underline{0.500} \\
 & Ours & 13 & \textbf{13} & \textbf{100} & \textbf{0.548} \\ \hline\hline
\multirow{6}{*}{\begin{tabular}[c]{@{}c@{}}\textit{HeLiPR}\\ (\texttt{Aeva3-Avia1})\end{tabular}} 
 & M2DP & 625 & 8 & 1.28 & 0.025 \\
 & SC & 625 & 48 & 7.68 & 0.142 \\
 & OT & 101 & 3 & 2.97 & 0.057 \\
 & LoGG3D & 28 & 1 & 3.57 & 0.039 \\
 & STD & 3 & \underline{3} & \textbf{100} & \underline{0.468} \\
 & Ours & 9 & \textbf{9} & \textbf{100} & \textbf{0.472} \\ \hline\hline
\multirow{6}{*}{\begin{tabular}[c]{@{}c@{}}\textit{INHA}\\ (\texttt{WHEEL-DOG})\end{tabular}} 
 & M2DP & 744 & 75 & 10.1 & 0.182 \\
 & SC & 12 & 12 & \textbf{100} & \textbf{0.709} \\
 & OT & 1 & 1 & 100.0 & 0.417 \\
 & LoGG3D & 50 & 2 & 4.0 & 0.041 \\
 & STD & 19 & \underline{19} & \textbf{100} & {0.500} \\
 & Ours & 31 & \textbf{31} & \textbf{100} & \underline{0.586} \\ \hline\hline
\multirow{6}{*}{\begin{tabular}[c]{@{}c@{}}\textit{INHA}\\ (\texttt{WHEEL-HAND1})\end{tabular}} 
 & M2DP & 466 & 37 & 7.94 & 0.144 \\
 & SC & 704 & 0 & 0 & 0.000 \\
 & OT & 8 & 1 & 12.5 & 0.167 \\
 & LoGG3D & 4 & 1 & 25.0 & 0.081 \\
 & STD & 2 & \underline{2} & \textbf{100} & \textbf{0.362} \\
 & Ours & 10 & \textbf{10} & \textbf{100} & \underline{0.356} \\ \hline\hline
\multirow{6}{*}{\begin{tabular}[c]{@{}c@{}}\textit{INHA}\\ (\texttt{DOG-HAND2})\end{tabular}} 
 & M2DP & 891 & 74 & 8.31 & 0.149 \\
 & SC & 912 & 105 & 11.5 & 0.202 \\
 & OT & 266 & 8 & 3.01 & 0.058 \\
 & LoGG3D & 812 & 9 & 1.11 & 0.021 \\
 & STD & 13 & 7 & \underline{53.8} & \underline{0.284} \\
 & Ours & 4 & \textbf{4} & \textbf{100} & \textbf{0.341} \\ \hline
\end{tabular}%
}
\vspace{-0.3cm}
\end{table}

From the perspective of multi-map merging, accurate but abundant loop pairs are crucial. Fig. \ref{fig:main_result} presents the matched loop pairs for maximum precision, and Table \ref{tab:match_pair} shows the number of total matching pairs, TP pairs at maximum precision, and F1 score across all sequences for each methodology. In Fig. \ref{fig:main_result}, green lines and red lines represent the true positive and false positive matching pairs for the top-1 match, respectively. The first row shows different types of LiDAR models for the central and query trajectories. As shown in Fig. \ref{fig:main_result}, our approach consistently exhibits a more dispersed set of true loop pairs compared to other algorithms. All methods except STD and our method show that most loop pairs are concentrated in a particular area, indicating that descriptors struggle to distinguish scenes precisely.

While STD shows robustness in LiDAR-modality, it is vulnerable in dynamic environments such as \textit{HeLiPR} or \textit{INHA}. 
In Fig. \ref{fig:main_result}, the dashed blue boxes in our method represent the locations identified as a loop, whereas the dashed orange boxes of STD display missing loop pair matches.

Removing dynamic objects (DynaSTD) enhanced performance by increasing the reliability of loop detection in dynamic environments, resulting in an 8.61\% improvement in the F1 score compared to STD.


\subsection{Map Merging and Alignment Evaluation}

We quantitatively evaluated the map alignment performance and compared our method (Uni-Mapper) to the SC-based LT-mapper (original), STD-based LT-Mapper (Customized), and the LTA-OM. 
Three sequence pairs are selected for evaluation: one from the \textit{HeLiPR} dataset (\texttt{Ouster1-Aeva2}) for multi-session scenarios, and two from the \textit{INHA} dataset (\texttt{WHEEL-DOG} and \texttt{WHEEL-HAND1}) for multi-robot systems, respectively. All pairs involve multi-modal LiDAR data.

\begin{figure*}[t!]
    \centering
	\def\width{2.0\columnwidth}%
	\includegraphics[width=\width, trim= 2.5cm 1cm 2.4cm 1cm, clip]{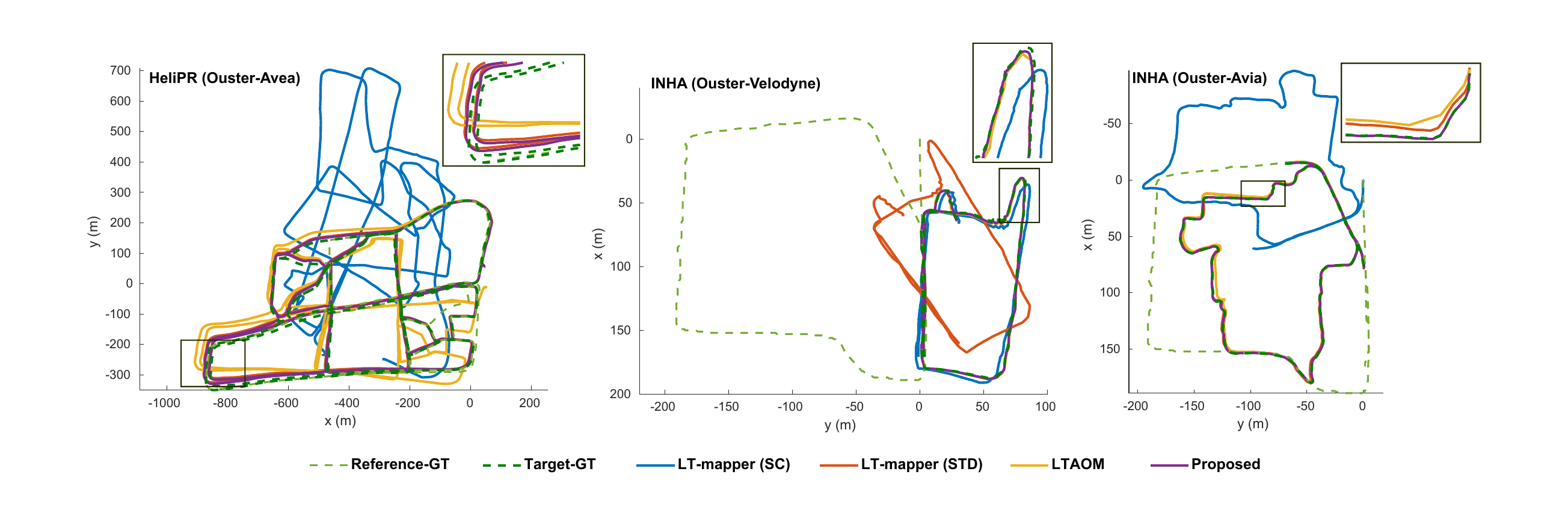}
 \captionsetup{font=footnotesize}
	\caption{Visualization of the trajectory after map alignment. Each trajectory represents the alignment results on \textit{HeLiPR} \texttt{(Ouster1-Avea2)}, \textit{INHA} \texttt{(WHEEL-DOG)}, and \textit{INHA} \texttt{(WHEEL-HAND1)} sequences. Also, the black box represents the enlarged view of the trajectories.}
    \vspace{-0.4cm}
    \label{fig:traj_result}
\end{figure*}

\begin{figure*}[t]
    \centering
	\def\width{1.8\columnwidth}%
	\includegraphics[width=\width, trim=0cm 7.0cm 0cm 0cm, clip]{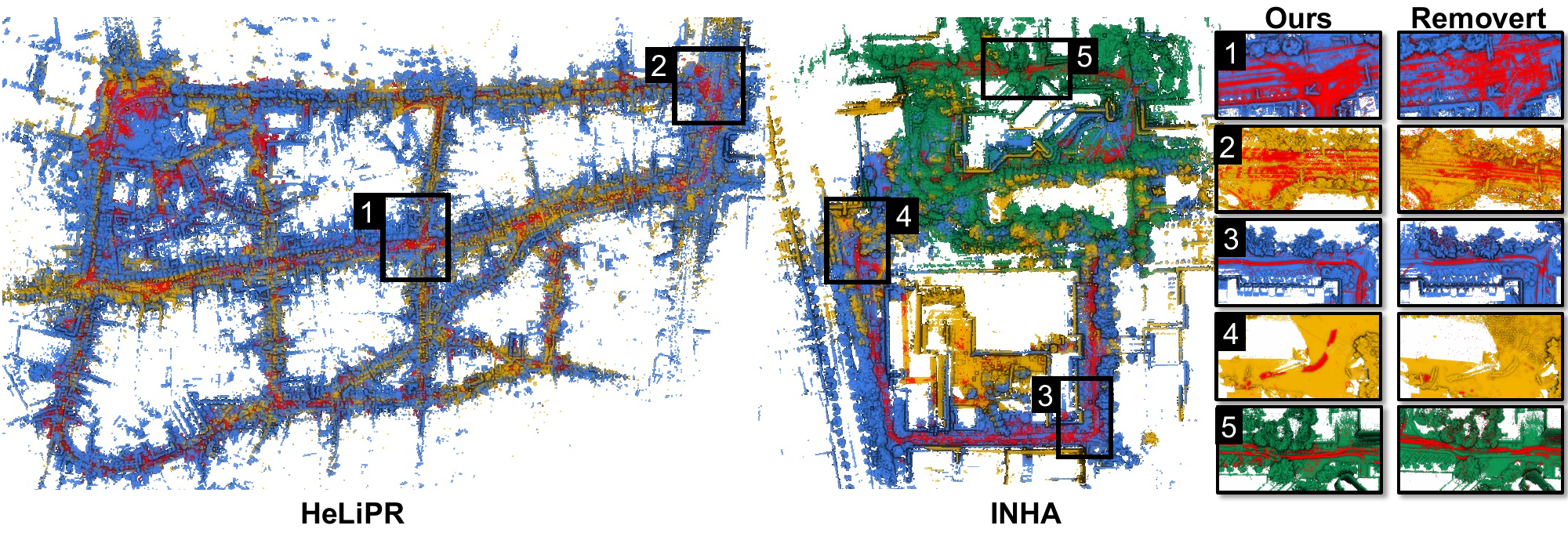}
 \captionsetup{font=footnotesize}
	\caption{Results of map merging and dynamic filtering. The point cloud maps represent merged maps between multi-modal LiDARs. The red points indicate dynamic objects. The pair of colors and sequence is represented as follows. \textit{HeLiPR}: \texttt{Ouster1} (blue), \texttt{Aeva2} (yellow) / \textit{INHA}: \texttt{WHEEL} (blue), \texttt{DOG} (green), and \texttt{HAND1} (yellow).} 
    \vspace{-0.4cm}
    \label{fig:merge_result}
\end{figure*}

Visualization of the optimized global trajectories and the unified map of each data sequence is illustrated in Fig. \ref{fig:traj_result} and Fig. \ref{fig:merge_result}. 
As the goal of map merging is to align the query map to the central map, we define the ground truth trajectory of the central map as Reference-GT and the ground truth trajectory of the query map, which should be globally aligned with Reference-GT, as Target-GT. 

\begin{table}[t]
\centering
\caption{Absolute Trajectory Error}
\label{tab:ape}
\resizebox{\columnwidth}{!}{%
\begin{tabular}{llllll}
\hline\hline
\multicolumn{2}{l}{} & LT-SC & LT-STD & LTA-OM & Uni-mapper \\ \hline\hline
\multirow{6}{*}{\begin{tabular}[c]{@{}c@{}}\textit{HeLiPR}\\ \texttt{(Ouster1-Aeva2)}\end{tabular}} 
 & max    & N/A & 46.0134        & N/A & \textbf{36.966} \\
 & mean   & N/A & 22.072         & N/A & \textbf{17.988} \\
 & median & N/A & 24.383         & N/A & \textbf{19.127} \\
 & min    & N/A & \textbf{0.287} & N/A & 0.438 \\
 & rmse   & N/A & 25.511         & N/A & \textbf{20.857} \\
 & std    & N/A & 12.791         & N/A & \textbf{10.558} \\ \hline
\multirow{6}{*}{\begin{tabular}[c]{@{}c@{}}\textit{INHA}\\ \texttt{(WHEEL-DOG)}\end{tabular}} 
 & max    & 12.099 & N/A & 2.196          & \textbf{1.338} \\
 & mean   & 6.905  & N/A & 1.224          & \textbf{0.817} \\
 & median & 7.918  & N/A & \textbf{0.995} & {1.059} \\
 & min    & 0.551  & N/A & 0.345          & \textbf{0.003} \\
 & rmse   & 7.663  & N/A & 1.397          & \textbf{0.951} \\
 & std    & 3.323  & N/A & 0.673          & \textbf{0.488} \\ \hline

\multirow{6}{*}{\begin{tabular}[c]{@{}c@{}}\textit{INHA}\\ \texttt{(WHEEL-DOG)}\end{tabular}} 
 & max    & N/A & 2.312 & 3.448 & \textbf{1.241} \\
 & mean   & N/A & 0.588 & 1.205 & \textbf{0.349} \\
 & median & N/A & 0.438 & 1.116 & \textbf{0.287} \\
 & min    & N/A & 0.180 & 0.175 & \textbf{0.115} \\
 & rmse   & N/A & 0.725 & 1.395 & \textbf{0.397} \\
 & std    & N/A & 0.424 & 0.702 & \textbf{0.189} \\ \hline
\end{tabular}%
}
\end{table}

Table \ref{tab:ape} represents the quantitative results of the map merging framework by ATE on three sequence pairs.
As ATE calculates the relative transformation between the ground truth pose and the estimated pose in global coordinates, it can measure the accuracy of map alignment. 
As demonstrated in Table \ref{tab:ape} and Fig. \ref{fig:traj_result}, our map-merging framework shows superior map alignment performance for dynamic environments and \ac{LiDAR} modalities.
As indicated in Sec. \ref{subsec:7-B}, LT-mapper fails to find a number of correct inter-map loop pairs in cross-modal LiDAR scenarios, such as \texttt{Ouster1-Aeva2} or \texttt{WHEEL-HAND1}. LT-mapper(STD) shows improved alignment performance in multi-modal scenarios. However, the \texttt{WHEEL-DOG} sequence pair, which has numerous pedestrians, fails to align two maps by false loop pairs and incorrect registration of local descriptors. 

While LTA-OM achieves map merging for \texttt{WHEEL-DOG} and \texttt{WHEEL-HAND1}, it fails to continually detect loops and associate data after initializing pose from the prior map in  the \textit{HeLiPR} sequence.

Our framework, utilizing the same parameter settings as STD, achieves robust performance even in highly dynamic environments by rejecting outliers and detecting more loop pairs than the original STD. 
In all scenarios, our framework exhibits the lowest \ac{RMSE}, attributable to our two-step optimization process. Fig. \ref{fig:merge} illustrates the total map merging process on \texttt{ Ouster1-Aeva2}. While the initial alignment of inter-map optimization via triangle descriptor significantly reduces errors, a residual z-axis drift persists. Through our refined registration process, we further minimize leftover drift errors, enabling more precise alignment.
It shows an 87\% improvement compared with the LT-mapper and 32\% compared with the LT-mapper(STD) in ATE criteria. 

\begin{table}[t]
\centering
\caption{Comparison of Mapping performance}
\label{tab:mapping_eval}
\resizebox{\columnwidth}{!}{%
\begin{tabular}{clccccc}
\hline\hline
 & Metric & LT-SC & LT-STD & LTA-OM & Ours \\ \hline\hline
\multirow{3}{*}{\begin{tabular}[c]{@{}c@{}}\textit{HeLiPR}\\ \texttt{(Ouster1-Aeva2)}\end{tabular}} 
 & AC[m] $\downarrow$    & N/A & 1.71 & N/A & \textbf{0.38} \\
 & CD[m] $\downarrow$    & N/A & 3.20 & N/A & \textbf{1.34} \\
 & MME $\downarrow$      & N/A & 3.02 & N/A & \textbf{2.91} \\ \hline
\multirow{3}{*}{\begin{tabular}[c]{@{}c@{}}\textit{INHA}\\ \texttt{(WHEEL-DOG)}\end{tabular}} 
 & AC[m] $\downarrow$    & 2.16 & N/A & 2.98 & \textbf{1.24} \\
 & CD[m] $\downarrow$    & 4.06 & N/A & 4.09 & \textbf{2.37} \\
 & MME $\downarrow$      & 3.12 & N/A & 3.08 & \textbf{3.04} \\ \hline
\multirow{3}{*}{\begin{tabular}[c]{@{}c@{}}\textit{INHA}\\ \texttt{(WHEEL-HAND1)}\end{tabular}} 
 & AC[m] $\downarrow$    & N/A & 2.31 & 2.40 & \textbf{1.25} \\
 & CD[m] $\downarrow$    & N/A & 4.03 & 4.18 & \textbf{2.73} \\
 & MME $\downarrow$      & N/A & 3.04 & 3.09 & \textbf{2.96} \\ \hline
\end{tabular}%
}
\vspace{-0.4cm}
\end{table}

\begin{figure}[h!]
    \centering
	\def\width{1.0\columnwidth}%
	\includegraphics[width=\width, trim=0 0 0 0, clip]{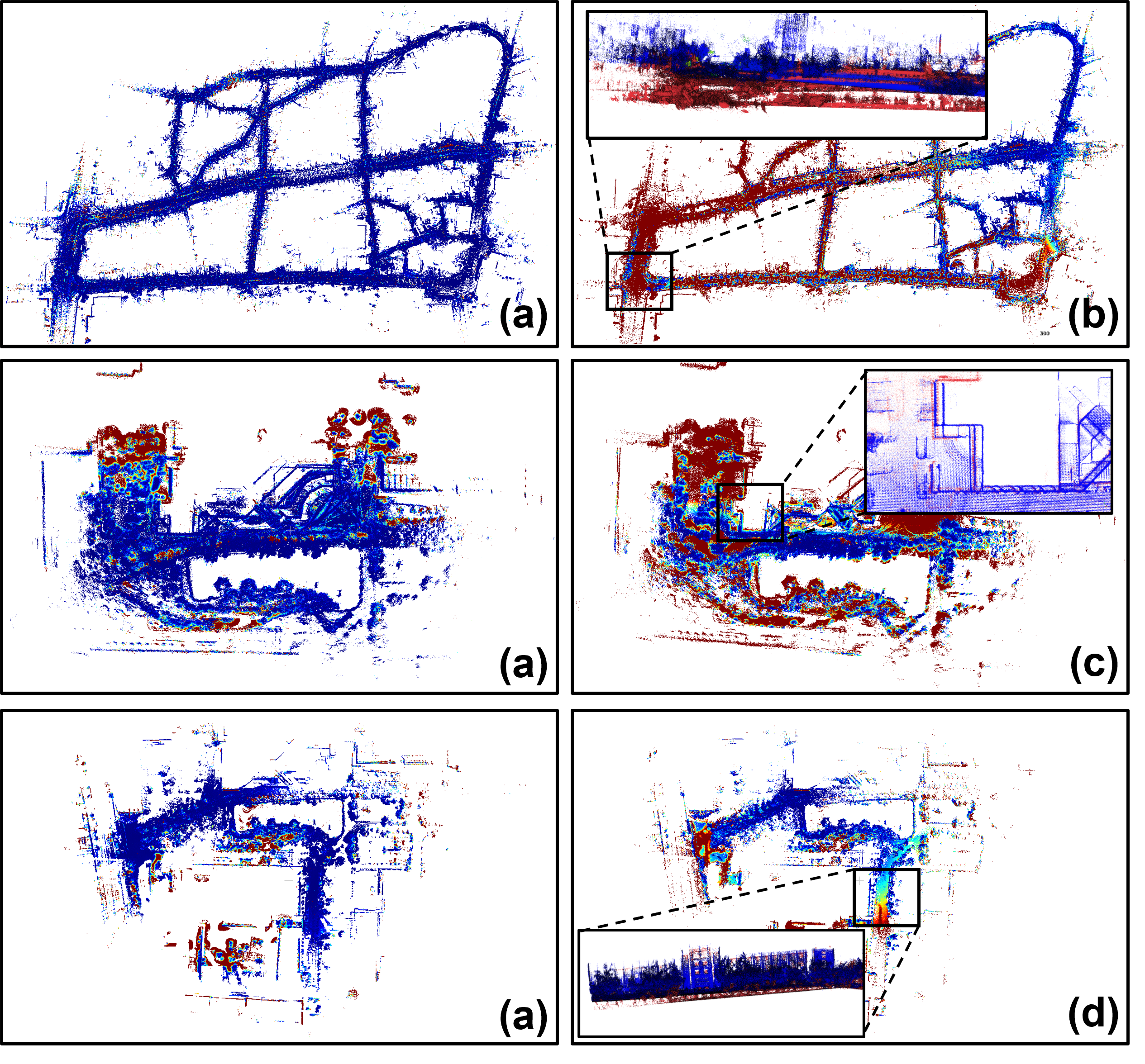}
 \captionsetup{font=footnotesize}
	\caption{Comparison of Chamfer Distance after map alignment: (a) Ours, (b) LT-STD, (c) LT-SC, (d) LTA-OM. The color gradient from blue to red indicates increasing distance errors, with blue for lower errors and red for higher errors. }
    \label{fig:Mapping}
    \vspace{-0.8cm}
\end{figure}

Since localization errors do not guarantee the mapping quality, we conducted additional experiments to evaluate the accuracy of the merged maps. Table. \ref{tab:mapping_eval} and Fig. \ref{fig:Mapping} present the detailed experimental results on the consistency of the merged point cloud maps. Across the three sequences, our framework demonstrates superior mapping accuracy compared to other map-merging systems based on AC, CD, and MME metrics. These results indicate that the point cloud maps from multiple sessions and robots are well-aligned, maintaining high consistency in the merged maps.
Fig. \ref{fig:Mapping} visualizes the CD results between our method and others for the two point cloud maps. Our method predominantly displays blue regions, indicating successful map registrations, while others show partial red areas, particularly where significant z-axis drift occurs, highlighting misalignment across multiple maps.

\begin{table}[t]
\centering
\caption{Ablation study on \textit{HeLiPR} dataset.}
\label{tab:ablation}
\begin{tabular}{lcccc}
\hline \hline
{Method} & {ATE [m] $\downarrow$} & {AC [m] $\downarrow$} & {CD [m] $\downarrow$} & {MME $\downarrow$}  \\ \hline \hline
w/o All         & 24.59 & 2.316   & 3.989   & 3.043      \\ 
w/o DOR         & 22.54 & 1.345   & 2.490   & 2.993      \\ 
w/o REG         & 24.69 & 2.166   & 3.531   & 3.028      \\ 
All             & \textbf{20.86} & \textbf{0.378}   & \textbf{1.336}   & \textbf{2.912}      \\ \hline
\end{tabular}
    \vspace{-0.5cm}
\end{table}

\begin{figure*}[]
    \centering
	\def\width{1.9\columnwidth}%
	\includegraphics[width=\width, trim=0cm 0cm 0cm 0cm, clip]{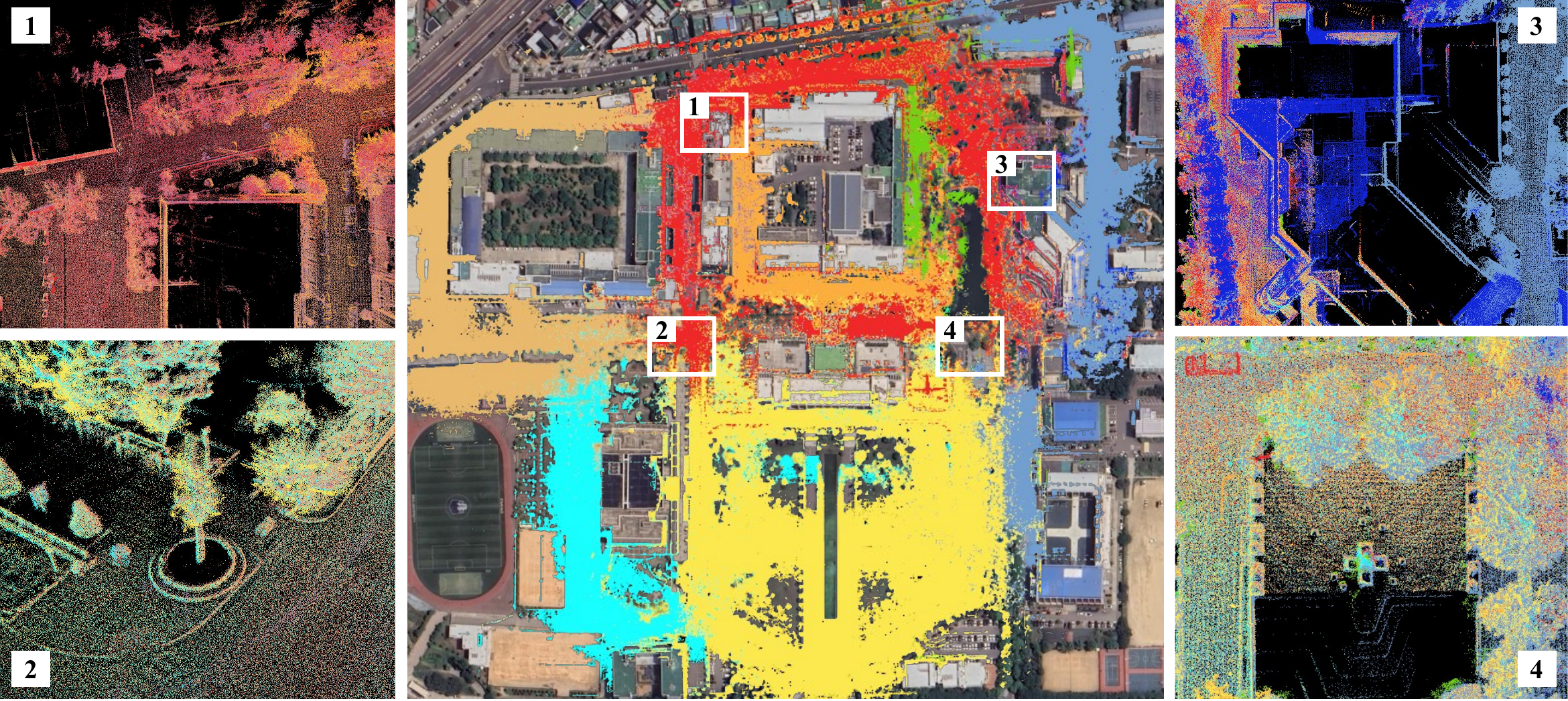}
    \captionsetup{font=footnotesize}
	\caption{
    Qualitative results of Uni-mapper for eight distinct maps on the INHA University campus. The red map acts as the central map, with others serving as query maps. The entire map merging process was conducted simultaneously, and only scan and pose information were provided.
    } 
    
    \label{fig:merged_maps}
\end{figure*}

\begin{figure*}[]
    \centering
	\def\width{1.9\columnwidth}%
	\includegraphics[width=\width, trim=0cm 0cm 0cm 0cm, clip]{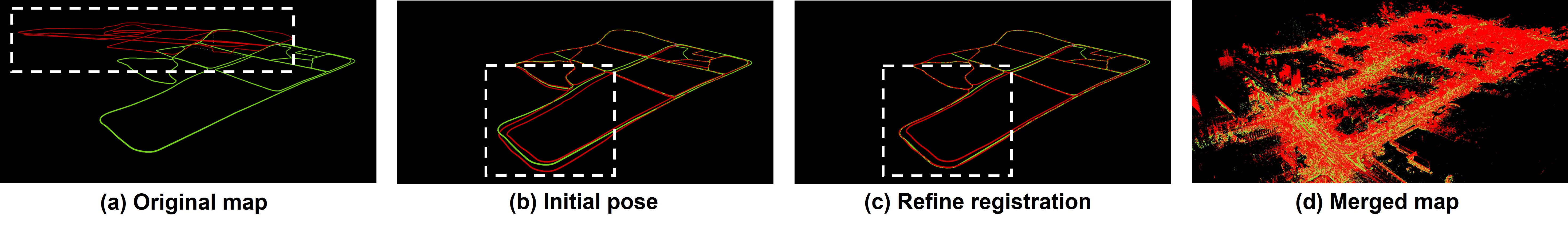}
 \captionsetup{font=footnotesize}
	\caption{Sequential process of 2-step registration for \textit{HeLiPR} \texttt{Ouster1-Aeva2} sequence. } 
    \vspace{-0.3cm}
    \label{fig:merge}
\end{figure*}

Table \ref{tab:ablation} presents the ablation studies of the submodules in our framework on the \textit{HeLiPR} sequence. Excluding all submodules leads to the worst performance in both ATE and mapping metrics (AC, CD, and MME). As shown in Table \ref{tab:ablation}, the 2-step registration (REG) achieves the greatest performance gap, highlighting the importance of refining pose errors and demonstrating that relative pose estimation based solely on scene descriptor is insufficient for consistent multi-map merging. Although the DOR module shows minimal impact overall, it becomes significant when loop pair detection fails, as seen in the \textit{INHA} \texttt{WHEEL-DOG} sequence. In such cases, without a successful initial pose estimation, the registration module has no opportunity to refine poses.

\begin{figure}[]
    \centering
	\def\width{0.9\columnwidth}%
	\includegraphics[width=\width, trim=0 0 0 0, clip]{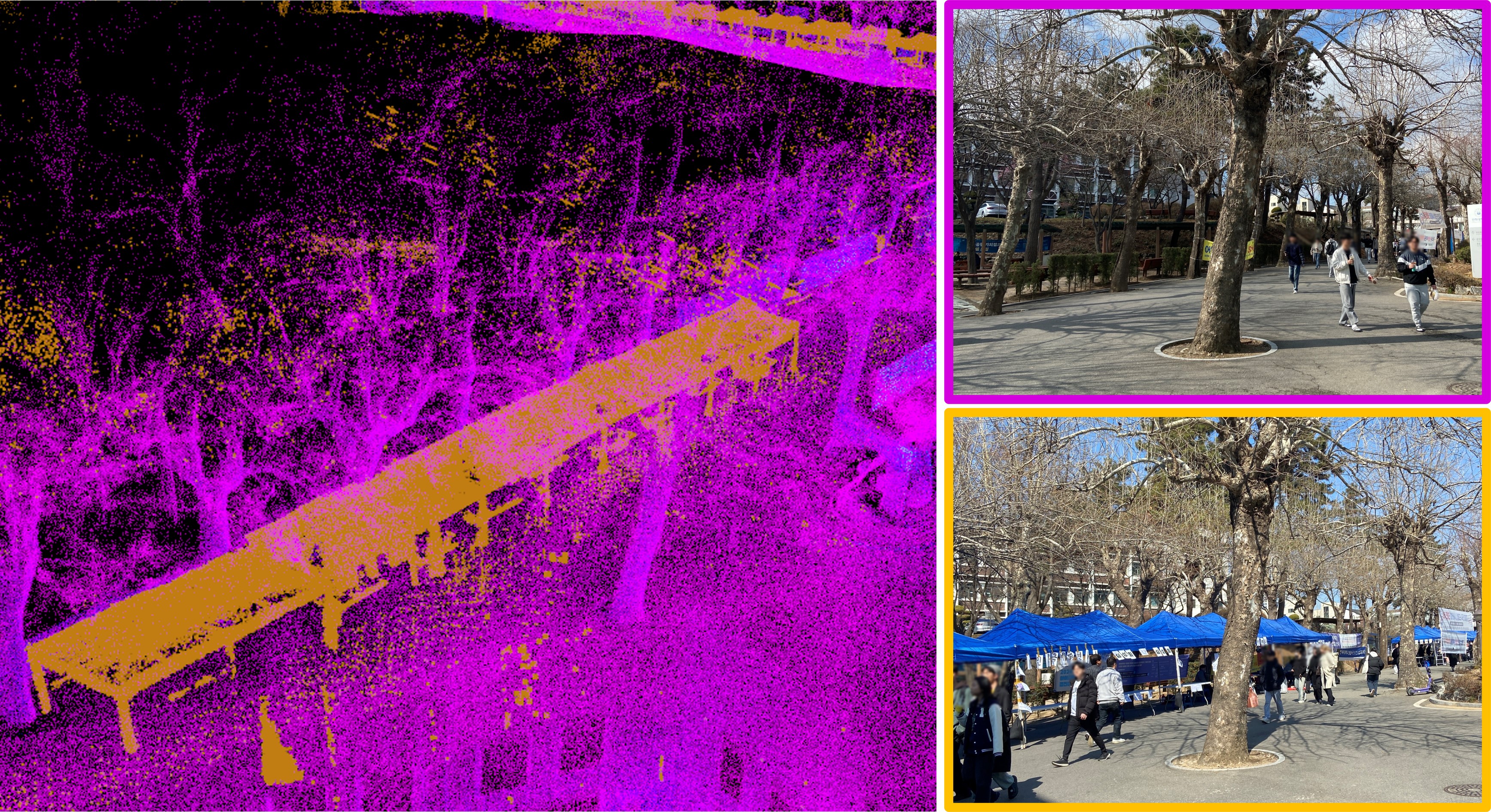}
 \captionsetup{font=footnotesize}
	\caption{Changed area from prior map (purple) to current map (orange). Inter-map alignment alleviates detecting changed areas.}
    \label{fig:CD}
    \vspace{-0.4cm}
\end{figure}

Additionally, our framework integrates multiple maps within a single process, enabling efficient crowd-sourced mapping without redundant remapping. As shown in Fig.\ref{fig:merged_maps} and Fig.\ref{fig:CD}, it aligns independently generated maps from diverse LiDAR sensors into a unified global map. This unified map provides well-aligned structures and supports scalable robotic autonomy by allowing efficient updates of only changed areas.



%% file: 9_Conclusion.tex
\section{Conclusion}

In this work, we introduce \textbf{Uni-Mapper}, a multi-map merging framework that integrates a dynamic-aware global descriptor in complex and dynamic environments.
The proposed dynamic removal module removes dynamic objects and ensures that the DynaSTD descriptor successfully captures cross-modal information from multi-modal \ac{LiDAR} sensors by utilizing a local keypoint combination of the coarse voxel. Finally, the Uni-Mapper framework integrates multiple maps through the optimization of the pose graph based on the anchor node and a two-step registration process. 
We validated the performance of DynaSTD and inter-robot loop closure representation of our framework with datasets, \textit{HeLiPR} and \textit{INHA}.
Compared with other \ac{SOTA} methods, Uni-Mapper demonstrates robust place recognition and outstanding map alignment by removing dynamic objects.

While the proposed framework effectively supports robust multi-map merging in dynamic environments, its performance remains sensitive to the quality of the central map.
When the central map contains sparse structures or geometric inconsistencies, the registration accuracy may deteriorate, leading to decreased overall map reliability and degraded local consistency across sessions.

To address this limitation, we plan to improve multi-session local consistency through LiDAR-based bundle adjustment. The framework is also extendable to a decentralized architecture for multi-agent \ac{SLAM}, enabling real-time data sharing among agents. In addition, we aim to enhance \ac{LiDAR} map management by detecting both temporal dynamics and long-term object-level changes, while integrating learning-based detection and segmentation to support high-level semantic understanding.